\documentclass[10pt,journal,compsoc]{IEEEtran}
\usepackage{amsmath,amsfonts}
\usepackage{algorithmic}
\usepackage{algorithm}
\usepackage{array}
\usepackage[caption=false,font=normalsize,labelfont=sf,textfont=sf]{subfig}
\usepackage{textcomp}
\usepackage{stfloats}
\usepackage{url}
\usepackage{verbatim}
\usepackage{graphicx}
\usepackage{hyperref}
\usepackage{url}
\usepackage{color}
\usepackage{subfloat}
\usepackage{graphicx}
\usepackage{multirow}
\usepackage{array}
\usepackage{booktabs}
\usepackage{ulem}
\usepackage{bm}
\usepackage{enumitem}
\usepackage{etoolbox}

\usepackage{amsthm}

\newtheoremstyle{bolddef}
  {.1em} 
  {.1em} 
  {} 
  {} 
  {\bfseries} 
  {.} 
  {.5em} 
  {} 

\theoremstyle{bolddef}
\newtheorem{definition}{Definition}
\newtheorem{proposition}{Proposition}

\ifCLASSOPTIONcompsoc
  \usepackage[nocompress]{cite}
\else
  \usepackage{cite}
\fi
\ifCLASSINFOpdf
\else
\fi
\begin{document}
%
\title{Energy-Latency Manipulation of Multi-modal Large Language Models via Verbose Samples}
%
%
%
%
\author{Kuofeng Gao\textsuperscript{*}, Jindong Gu\textsuperscript{*}, Yang Bai, Shu-Tao Xia\textsuperscript{$\dagger$}, Philip Torr, \textit{Senior Member, IEEE},\\  Wei Liu, \textit{Fellow, IEEE}, Zhifeng Li\textsuperscript{$\dagger$}, \textit{Senior Member, IEEE}
\IEEEcompsocitemizethanks{
\IEEEcompsocthanksitem * Equal contribution.
\IEEEcompsocthanksitem
Kuofeng Gao and Shu-Tao Xia are with Tsinghua Shenzhen International Graduate School, Tsinghua University, Shenzhen, Guangdong, China and Shu-Tao Xia is also with the Peng Cheng Laboratory, Shenzhen, Guangdong, China.
(E-mail: gkf21@mails.tsinghua.edu.cn,  xiast@sz.tsinghua.edu.cn).
\IEEEcompsocthanksitem Jindong Gu and Philip Torr are with Torr Vision Group, University of Oxford. (E-mail: jindong.gu@eng.ox.ac.uk, philip.torr@eng.ox.ac.uk)
\IEEEcompsocthanksitem Yang Bai is with Tencent Technology (Beijing) Co.Ltd, Beijing, China. (E-mail: mavisbai@tencent.com)
\IEEEcompsocthanksitem Wei Liu and Zhifeng Li are with Tencent Data Platform, ShenZhen, China. (E-mail: wl2223@columbia.edu, michaelzfli@tencent.com)
\IEEEcompsocthanksitem \textsuperscript{$\dagger$}Corresponding authors: Shu-Tao Xia (E-mail: xiast@sz.tsinghua.edu.cn) and Zhifeng Li (E-mail: michaelzfli@tencent.com).
}
}

\markboth{Gao and Gu and Bai \MakeLowercase{\textit{et al.}}: Energy-Latency Manipulation of Multi-modal Large Language Models via Verbose Samples}
{Shell \MakeLowercase{\textit{et al.}}: A Sample Article Using IEEEtran.cls for IEEE Journals}

\IEEEtitleabstractindextext{%
\begin{abstract}
Despite the exceptional performance of multi-modal large language models (MLLMs), their deployment requires substantial computational resources. Once malicious users induce high energy consumption and latency time (energy-latency cost), it will exhaust computational resources and harm availability of service. In this paper, we investigate this vulnerability for MLLMs, particularly image-based and video-based ones, and aim to induce high energy-latency cost during inference by crafting an imperceptible perturbation. We find that high energy-latency cost can be manipulated by maximizing the length of generated sequences, which motivates us to propose \textit{verbose samples}, including \textit{verbose images and videos}. Concretely, two modality non-specific losses are proposed, including a loss to delay end-of-sequence (EOS) token and an uncertainty loss to increase the uncertainty over each generated token. In addition, improving diversity is important to encourage longer responses by increasing the complexity, which inspires the following modality specific loss. For verbose images, a token diversity loss is proposed to promote diverse hidden states. For verbose videos, a frame feature diversity loss is proposed to increase the feature diversity among frames. To balance these losses, we propose a temporal weight adjustment algorithm. Experiments demonstrate that our verbose samples can largely extend the length of generated sequences. 
\end{abstract}

\begin{IEEEkeywords}
Energy-latency manipulation, multi-modal large language models, verbose samples.
\end{IEEEkeywords}}

\maketitle

\IEEEdisplaynontitleabstractindextext

%
\IEEEpeerreviewmaketitle

\IEEEraisesectionheading{\section{Introduction}}
\label{sec:intro}
\IEEEPARstart{M}{ulti-modal} large language models (MLLMs) \cite{alayrac2022flamingo,chen2022visualgpt,liu2023visual,li2021align,li2023blip,openai2023gpt4}, including image-based LLMs \cite{li2022blip,dai2023instructblip,zhu2023minigpt,peng2023kosmos} and video-based LLMs \cite{li2023mvbench,zhang2023video,lin2023video},  have achieved remarkable performance across various multi-modal tasks in both image and video modality, including image and video captioning, question answering, and comprehension. However, these MLLMs \cite{openai2023gpt4,chen2023shikra,chen2023minigpt,li2023mvbench,maaz2023video} often consist of billions of parameters, which require significant computational resources for deployment \cite{bai2024beyond}. The larger scale of these models can lead to increased energy consumption and longer latency time during the inference process. Both NVIDIA and Amazon Web Services \cite{patterson2021carbon} have stated that the deployment for the inference process accounts for over 90\% of machine learning demand. This high demand for computational resources not only poses challenges in terms of cost but also raises concerns about the potential security risks.

Once attackers maliciously induce high energy consumption and latency time (energy-latency cost) during the inference stage, it can exhaust computational resources and reduce the availability of MLLMs.
The energy consumption is the amount of energy used on hardware during an inference, while the latency time represents the response time taken for the inference.
Previous studies have discussed how to manipulate high energy-latency cost in other architectures and specifically proposed sponge samples \cite{shumailov2021sponge} for LLMs, and NICGSlowdown \cite{chen2022nicgslowdown} for smaller-scale models. However, when these methods are directly applied to MLLMs, there remains room for improvement, which will be further discussed in Section~\ref{sec:related}. Different from LLMs and smaller-scale models, MLLMs integrate the image and video modality into impressive LLMs \cite{touvron2023llama,chowdhery2022palm} to enable powerful visual interaction with images and videos. This integration introduces vulnerabilities from the manipulation of visual inputs \cite{goodfellow2014explaining}, requiring specific methods to induce high energy-latency cost.

\begin{figure*}[t]
\begin{minipage}{\textwidth}
\centering
\subfloat[Energy of MiniGPT-4] {   
\label{InstructBLIP of linear correlation of energy and latency}  \includegraphics[width=0.23\textwidth]{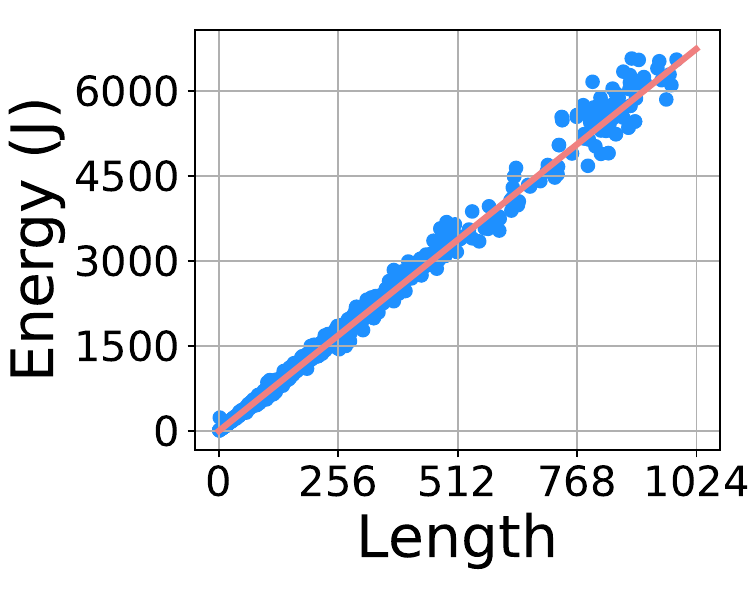} 
}    
\subfloat[Latency of MiniGPT-4] {   
\label{MiniGPT-4 of linear correlation of energy and latency}  \includegraphics[width=0.23\textwidth]{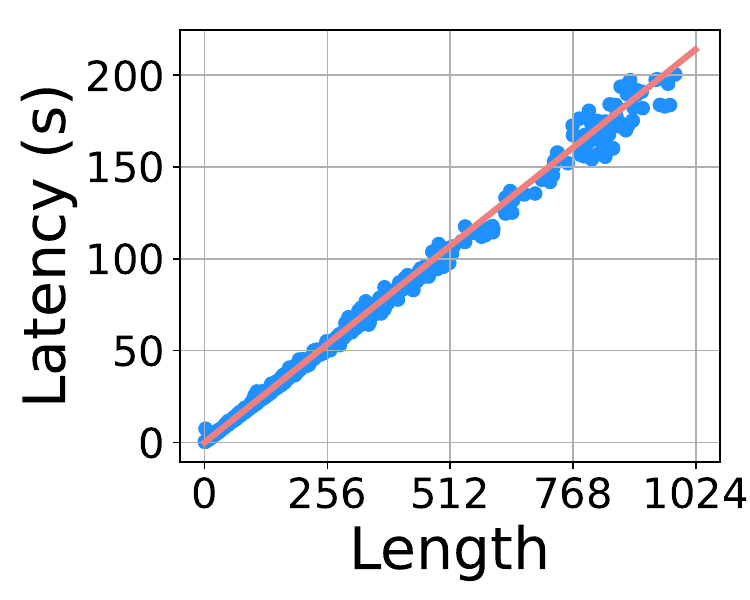} 
}    
\subfloat[Energy of VideoChat-2] {
\label{VideoChat of linear correlation of energy}  
\includegraphics[width=0.23\textwidth]{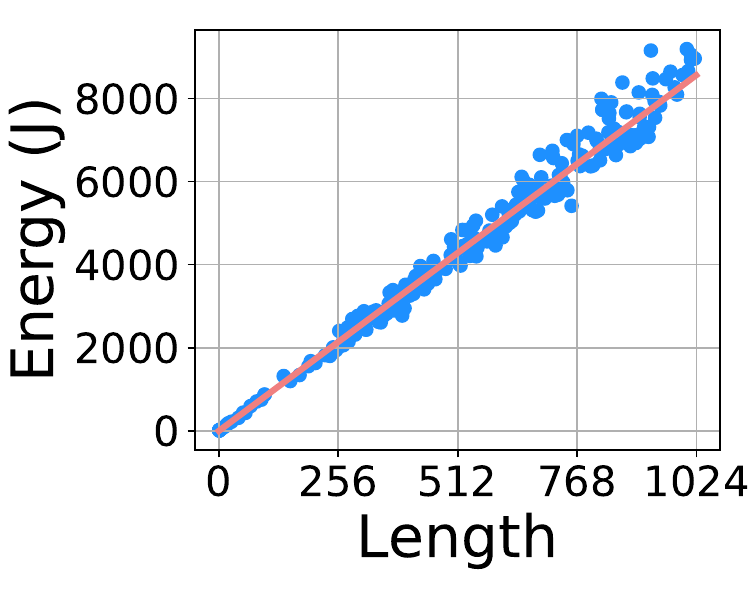}  
}        
\subfloat[Latency of VideoChat-2] {   
\label{VideoChat of linear correlation of latency}  \includegraphics[width=0.23\textwidth]{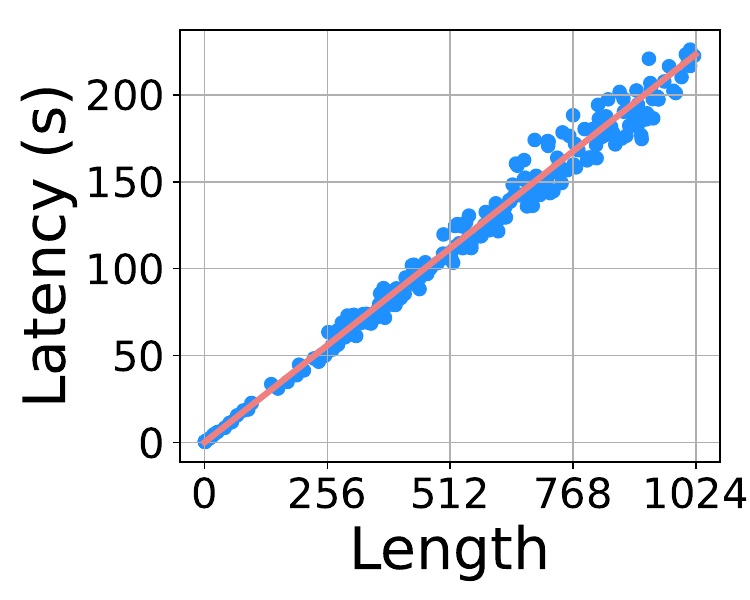} 
}    
\caption{The approximately positive linear relationship between energy consumption, latency time, and the length of generated sequences in image-based LLMs and video-based LLMs. Following \cite{shumailov2021sponge}, energy consumption is estimated by NVIDIA Management Library (NVML), and latency time is the response time of an inference. } 
\label{all linear correlation of energy and latency} 
\vspace{-1em}
\end{minipage}
\end{figure*}

In this paper, we first conduct a comprehensive investigation on energy consumption, latency time, and the length of generated sequences by MLLMs during the inference stage. 
As observed in Fig.~\ref{all linear correlation of energy and latency}, both energy consumption and latency time exhibit an approximately positive linear relationship with the length of generated sequences.  
Hence, the high energy-latency cost of MLLMs can be maximized by increasing the length of their generated sequences. 
Consequently, we propose \textbf{\textit{verbose samples}}, including \textbf{\textit{verbose images}} and \textbf{\textit{verbose videos}}, to craft an imperceptible perturbation, which can induce image-based LLMs and video-based LLMs to generate long sentences during the inference, respectively.

For both verbose images and videos, two modality non-specific loss objectives are designed as follows. (1) \textbf{Delayed EOS loss}: The end-of-sequence (EOS) token is a signal to stop generating further tokens. By delaying the placement of the EOS token, MLLMs are encouraged to generate more tokens and extend the length of generated sequences. Besides, to accelerate the process of delaying EOS tokens, we propose to break output dependency, following \cite{chen2022nicgslowdown}. With the guide of the Delayed EOS loss and diverging from the original output dependency, the generated sequence can be deviated towards the direction of a longer sequence. (2) \textbf{Uncertainty Loss}: By introducing more uncertainty over each generated token, it can break the original output dependency at the token level and encourage MLLMs to produce more varied outputs and generate longer sequences.

Additionally, to encourage longer generated sequences, we propose to improve diversity during inference, which can increase the complexity of the output. Based on different input modalities, the modality-specific loss objective for each modality is designed as follows. For verbose images, we propose a \textbf{Token Diversity Loss} that encourages token diversity among all tokens of the whole generated sequence. As a result, it can break the original output dependency at the sequence level and enable image-based LLMs to generate a diverse range of tokens in the output sequence, contributing to longer and more complex sequences. 
For verbose videos, we propose a \textbf{Frame Feature Diversity Loss} tailored for video-based LLMs. Typically, a video consists of a series of consecutive frames representing a scene, with adjacent frames often sharing similar semantics. Leveraging this consistent semantic relationship, video-based LLMs can aggregate different frames within a video into features to produce a coherent sequence. Hence, to promote a longer generated sequence, we propose to increase the diversity of frame features, which can induce different frames away from each other and introduce inconsistent semantics. Consequently, it can confuse video-based LLMs, making it hard to consistently comprehend the content in videos.

Both verbose images and videos introduce three loss objectives. To balance them, we propose a temporal weight adjustment algorithm for the optimization of verbose images and videos. Extensive experiments demonstrate that our verbose images can increase the length of generated sequences of image-based LLMs by 7.87$\times$ and 8.56$\times$ compared to original images on MS-COCO and ImageNet datasets. Our verbose videos can increase the length of generated sequences of video-based LLMs by 4.04$\times$ and 4.14$\times$ compared to original videos on MSVD and TGIF datasets.

In summary, a preliminary version of this work has been presented as the ICLR 2024 conference \cite{gao2024inducing} and contributions in the conference version can be outlined as follows:
\begin{itemize}
\item We conduct a comprehensive investigation and observe that energy consumption and latency time are approximately positively linearly correlated with the length of generated sequences for image-based LLMs. 

\item We propose \textbf{\textit{verbose images}} to craft an imperceptible perturbation to induce high energy-latency cost for image-based LLMs, which is achieved by delaying the EOS token, enhancing output uncertainty, improving token diversity, and employing a temporal weight adjustment algorithm during the optimization process.

\item Experiments show that our verbose images can increase the maximum length of generated sequences on MS-COCO and ImageNet across four image-based LLMs. Additionally, our verbose images can generate complex sequences containing hallucinated contents and produce dispersed attention on visual input.

\end{itemize}

In the current work, we incorporate the following additional contributions:
\begin{itemize}
\item For energy-latency manipulation of MLLMs, including both image-based LLMs and video-based LLMs, we find a general observation that energy consumption and latency time are approximately positively linearly correlated with the length of generated sequences. 

\item To induce high energy-latency cost for MLLMs, we propose \textbf{\textit{verbose samples}}, including \textbf{\textit{verbose images and videos}}, which both adopt delaying the EOS token and enhancing output uncertainty. Besides, we propose to improve token diversity for verbose images and improve frame feature diversity for verbose videos. Furthermore, a temporal weight adjustment algorithm is proposed to balance these losses. 

\item Experiments demonstrate that our verbose samples can generate the longest sequences. Besides, to highlight the necessity of verbose videos tailored for video-based LLMs, we compare and discuss the differences between verbose images and videos. Furthermore, we unify an interpretation framework for energy-latency manipulation in both image and video modality from hallucination and visual attention.

\item We supplement all discussions about the effect of both verbose images and videos, including different perturbation magnitudes, different adversary knowledge, and different multi-modal tasks.

\end{itemize}

\section{Related Work}
\label{sec:related}

\subsection{Multi-modal large language models (MLLMs)}
Recently, the advanced MLLMs, including image-based LLMs and video-based LLMs, have achieved an enhanced zero-shot performance in various multi-modal tasks \cite{silberer2016visually,peng2020mra,zhang2023universal}. 
The integration of the image or video modality into MLLMs enables visual context-aware interaction, surpassing the capabilities of existing LLMs. However, this integration also introduces vulnerabilities arising from the manipulation of visual inputs, such as adversarial misdirection \cite{zhao2023evaluating,dong2023robust,carlini2023aligned,gao2024adversarial,bai2023badclip,luo2024image,wang2024transferable} and visual jailbreak \cite{qi2023visual,wu2023jailbreaking,li2024images}. With the increasing size of LLM parameters and the inclusion of vision encoders, MLLMs consist of larger parameters, which results in higher energy-latency cost during the inference stage. In our paper, we aim to uncover this security threat in energy-latency manipulation of MLLMs during deployment. To this end, we propose to generate verbose samples to induce high energy-latency cost of MLLMs.

\subsection{Energy-latency manipulation}
The energy-latency manipulation \cite{chen2022nmtsloth,hong2020panda,chen2023dark,liu2023slowlidar} aims to slow down the models by increasing their energy computation and response time during the inference stage, a threat analogous to the denial-of-service (DoS) attacks \cite{pelechrinis2010denial} from the Internet. 
Concretely, sponge samples \cite{shumailov2021sponge} first observe that a larger representation dimension calculation can introduce more energy-latency cost in LLMs. Hence, they propose to craft sponge samples to maximize the $\mathcal{L}_2$ norm of activation values across all layers, thereby introducing more representation calculation and energy-latency cost.
NICGSlowDown \cite{chen2022nicgslowdown} proposes to increase the number of decoder calls, \textit{i.e.}, the length of the generated sequence, to increase the energy-latency of smaller-scale captioning models. They minimize the logits of both EOS token and output tokens to generate long sentences.

However, there is still room for improvement if these previous methods are directly applied to MLLMs for two main reasons. On one hand, they primarily focus on LLMs or smaller-scale models. Sponge samples are designed for LLMs for translations \cite{liu2019roberta} and NICGSlowdown targets for RNNs or LSTMs combined with CNNs for image captioning \cite{anderson2018bottom}. Differently, our verbose samples are tailored for MLLMs in multi-modal tasks. On the other hand, the objective of NICGSlowdown involves logits of specific output tokens. Nevertheless, current MLLMs generate random output sequences for the same input sample, due to advanced sampling policies \cite{holtzman2019curious}, which makes it challenging to optimize objectives with specific output tokens. Therefore, it highlights the need for methods specifically designed for MLLMs to induce high energy-latency cost.

\section{Methodology}
\subsection{Threat model}

\textbf{Goals and capabilities.} The goal of our proposed verbose samples is to craft an imperceptible perturbation and induce the MLLMs to generate a sequence as long as possible, thereby increasing the energy consumption and prolonging latency during the victim model's deployment. Specifically, the involved perturbation is restricted within a predefined magnitude in $l_p$ norm, ensuring that it is imperceptible and difficult to detect. 

\textbf{Knowledge and background.} 
We consider the target MLLMs which generate sequences using an auto-regressive process. As suggested in~\cite{bagdasaryan2023ab,qi2023visual}, we assume that the victim MLLMs can be accessed in full knowledge, including architectures and parameters, in our main experiments.  Additionally, we consider a more challenging scenario where the victim MLLMs are inaccessible in discussions. In such cases, the verbose samples can only be generated on surrogate MLLMs and subsequently transferred to the victim MLLMs.


\subsection{Problem formulation}

Consider an input image $\bm{x}$ for image-based LLMs and an input video with $M$ frames $\bm{X}=\{\bm{X}_1, ..., \bm{X}_M\}$ for video-based LLMs. They are accordingly with an input text $\bm{c}_{\text{in}}$ and a sequence of generated output tokens $\bm{y}=\{y_1, y_2, ..., y_N\}$, where $y_i$ represents the $i$-th generated token, $N$ is the length of the output sequence and $\bm{c}_{\text{in}}$ is a placeholder $\emptyset$ in image and video captioning or a question in image and video question answering. 
Based on the probability distribution over generated tokens, MLLMs generate one token at one time in an auto-regressive manner.
The probability distribution after the $\operatorname{Softmax}(\cdot)$ layer over the $i$-th generated token can be denoted as $f_i\left(y_1, \cdots, y_{i-1};\ \bm{x};\ \bm{c}_{\text{in}}\right)$ for image-based LLMs and $F_i\left(y_1, \cdots, y_{i-1};\ \bm{X};\ \bm{c}_{\text{in}}\right)$ for video-based LLMs. Since we mainly focus on visual inputs of MLLMs in this paper, we abbreviate it as $f_i\left(\bm{x}\right)$ and $F_i\left(\bm{X}\right)$, where $f_i\left(\bm{x}\right) \in \mathbb{R}^{\mathrm{V}}$, $F_i\left(\bm{X}\right) \in \mathbb{R}^{\mathrm{V}}$, and $\mathrm{V}$ is the vocabulary size. Meanwhile, the hidden states across all the layers over the $i$-th generated token are recorded as $g_i\left(y_1, \cdots, y_{i-1};\ \bm{x};\ \bm{c}_{\text{in}}\right)$, abbreviated as  $g_i\left(\bm{x}\right)$, where $g_i\left(\bm{x}\right) \in \mathbb{R}^{\mathrm{C}}$ and $\mathrm{C}$ is the dimension size of hidden states for image-based LLMs. In addition, the frame feature over the $j$-th frame can be extracted by $h_j\left(\bm{X}\right)$ for video-based LLMs, where $h_j\left(\bm{X}\right) \in \mathbb{R}^{\mathrm{D}}$ and $\mathrm{D}$ is the dimension size of frame features.


As discussed in Section \ref{sec:intro}, the energy consumption and latency time of an inference are approximately positively linearly related to the length of the generated sequence. Hence, inducing high energy-latency cost can be formulated as maximizing the length $N$ of the output tokens of MLLMs by crafting verbose images $\bm{x}'$ and verbose videos $\bm{X}'$. To ensure the imperceptibility, we impose an $l_p$ restriction on the imperceptible perturbations, where the perturbation magnitude is denoted as $\epsilon$, such that $||\bm{x}'-\bm{x}||_{p} \le \epsilon$ and $\frac{1}{M}\sum_{j=1}^M||\bm{X}_j'-\bm{X}_j||_{p} \le \epsilon$.





\begin{figure*}[t] \centering    
\includegraphics[width=\textwidth]{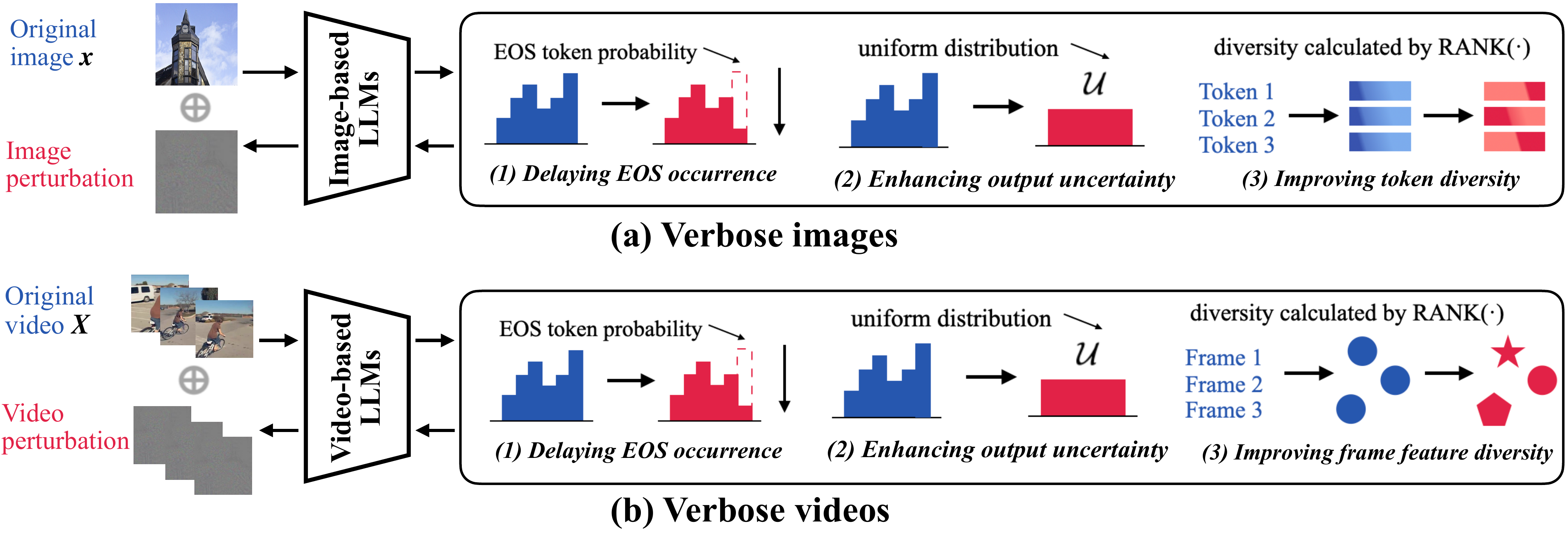}  
\vspace{-2em}
\caption{An overview of verbose samples against MLLMs to increase the length of generated sequences, thereby inducing higher energy-latency cost. Two modality non-specific losses are designed to delay EOS occurrence and enhance output uncertainty. Besides, a modality specific loss is proposed for each modality. For verbose images, the goal is to improve token diversity, while for verbose videos, is to improve frame feature diversity. Moreover, a temporal weight adjustment algorithm is proposed to better utilize the three objectives.} 
\label{simple illustration}
\vspace{-1em}
\end{figure*}

\subsection{Overview}

To increase the length of generated sequences, we propose two modality non-specific loss objectives for both verbose images and videos and the modality specific loss objective for each modality to optimize imperceptible perturbations. Two modality non-specific loss objectives are designed as follows. Firstly and straightforwardly, we propose a \textbf{delayed EOS loss} to hinder the occurrence of EOS token and thus force the sentence to continue.
However, the auto-regressive textual generation in MLLMs establishes an output dependency, which means that the current token is generated based on all previously generated tokens. 
Hence, when previously generated tokens remain unchanged, it is also hard to generate a longer sequence even though the probability of the EOS token has been minimized. 
To this end, we propose to break this output dependency as suggested in \cite{chen2022nicgslowdown}.
Concretely, we propose an \textbf{uncertainty loss} to enhance output uncertainty over each generated token.
Subsequently, the modality specific loss objective for each modality is designed as follows. We find that it can lead to the longer generated sequences by improving diversity, which can increase the complexity of the output.
For verbose images, we propose a \textbf{token diversity loss}, which improves the diversity among all tokens of the whole generated sequence. 
For verbose videos, we propose a \textbf{frame feature diversity loss} to increase the diversity of frame features within an input video.
Moreover, to balance three loss objectives during the optimization, a temporal weight adjustment algorithm is introduced in Section 
\ref{sec:loss weight}. Fig. \ref{simple illustration} shows an overview of our samples.

\subsection{Loss design}
\subsubsection{Delaying EOS occurrence}
\label{sec:loss design}

For both MLLMs, the auto-regressive generation process continues until an end-of-sequence (EOS) token is generated or a predefined maximum token length is reached. To increase the length of generated sequences, one straightforward approach is to prevent the occurrence of the EOS token during the prediction process. However, considering that the auto-regressive prediction is a non-deterministic random process, it is challenging to directly determine the exact location of the EOS token occurrence. 
Therefore, we propose to minimize the probability of the EOS token at all positions. This can be achieved through the delayed EOS loss, formulated as follows for verbose images:
\begin{equation}
\begin{aligned}
\mathcal{L}_{1}(\bm{x}')=\frac{1}{N} \sum_{i=1}^N f_{i}^{\mathrm{EOS}}\left(\bm{x}'\right),
\end{aligned}
\label{eq:eos loss}
\end{equation}
and formulated as follows for verbose videos:
\begin{equation}
\begin{aligned}
\mathcal{L}_{1}(\bm{X}')=\frac{1}{N} \sum_{i=1}^N F_{i}^{\mathrm{EOS}}\left(\bm{X}'\right),
\end{aligned}
\label{eq:eos loss videos}
\end{equation}
where $f_{i}^{\mathrm{EOS}}(\cdot)$ and $F_{i}^{\mathrm{EOS}}(\cdot)$ are the EOS token probability of the probability distribution after the $\operatorname{Softmax}(\cdot)$ layer over the $i$-th generated token in image-based LLMs and video-based LLMs, respectively.
When reducing the likelihood of every EOS token occurring by minimizing $\mathcal{L}_{1}(\cdot)$, MLLMs are encouraged to generate more and more tokens before reaching the EOS token.

\subsubsection{Enhancing output uncertainty}
To generate a longer sequence, we propose to break output dependency. This is because guided by the delay of the EOS token, the generated sequence can deviate towards a longer generated sequence, diverging from the original auto-regressive output dependency. Then, the problem can be formulated as how to induce the generated sequence away from the original sequence. Typically, MLLMs generate each token with the highest probability, forming a generated sequence. The probability of the generated sequence can be calculated as a product of conditional probabilities of each generated token \cite{bengio2000neural}, formulated as follows for verbose images:
\begin{equation}
\begin{aligned}
P(y_1...y_N;\bm{x};c_{\text{in}})=\prod \limits_{i=1}^N P(y_i|y_1,...,y_{i-1};\bm{x};c_{\text{in}}),
\end{aligned}
\label{eq:LLM generation}
\end{equation}
and formulated as follows for verbose videos:
\begin{equation}
\begin{aligned}
P(y_1...y_N;\bm{X};c_{\text{in}})=\prod \limits_{i=1}^N P(y_i|y_1,...,y_{i-1};\bm{X};c_{\text{in}}).
\end{aligned}
\label{eq:LLM generation videos}
\end{equation}
Consequently, the probability of the original generated sequence is also the highest probability. To deviate from the original generated sequence, we can reduce the probability of the generated sequence, which can be achieved by lowering the probability of each generated token. 
To this end, we propose to enhance output uncertainty over each generated token to facilitate longer and more complex output sequences. 
This objective can be implemented by maximizing the entropy of the output probability distribution for each generated token. Based on \cite{shannon1948mathematical}, it can be converted to minimize the Kullback–Leibler ($\mathrm{KL}$) divergence $D_{\mathrm{KL}}(\cdot, \cdot)$ \cite{kullback1951information} between the output probability distribution and a uniform distribution $\mathcal{U}$. 
Therefore, the uncertainty loss can be formulated as follows for verbose images:
\begin{equation}
\begin{aligned}
\mathcal{L}_{2}(\bm{x}')=\sum_{i=1}^{N} D_{\mathrm{KL}}\left(f_{i}\left(\bm{x}'\right), \mathcal{U}\right),
\end{aligned}
\label{eq:uncertainty loss}
\end{equation}
and formulated as follows for verbose videos:
\begin{equation}
\begin{aligned}
\mathcal{L}_{2}(\bm{X}')=\sum_{i=1}^{N} D_{\mathrm{KL}}\left(f_{i}\left(\bm{X}'\right), \mathcal{U}\right),
\end{aligned}
\label{eq:uncertainty loss videos}
\end{equation}
where $f_{i}(\cdot)$ and $F_{i}(\cdot)$ are the probability distribution after the $\operatorname{Softmax}(\cdot)$ layer over the $i$-th generated token in image-based LLMs and video-based LLMs, respectively. The uncertainty loss can introduce more uncertainty in the prediction for each generated token, effectively breaking the original output dependency. Consequently, when the original output dependency is disrupted,  MLLMs can generate more complex sentences and longer generated sequences, guided by the delay of the EOS token.

\subsubsection{Improving diversity for each modality}
For verbose images, to increase the complexity of the output, we propose to improve the diversity of hidden states among all generated tokens, which explores a wider range of possible outputs. Specifically, the hidden state of a token is the vector representation of a word or subword in image-based LLMs.

\begin{definition}
\label{definition}
\textit{Let $\operatorname{Rank}(\cdot)$ indicates the rank of a matrix and $[g_1(\bm{x}');g_2(\bm{x}');\cdots;g_N(\bm{x}')]$ denotes the concatenated matrix of hidden states among all generated tokens. To induce high energy-latency cost of image-based LLMs, the token diversity is defined as the rank of hidden states among all generated tokens, \textit{i.e.}, $\operatorname{Rank}([g_1(\bm{x}');g_2(\bm{x}');\cdots;g_N(\bm{x}')])$.}
\end{definition}

Given by Definition \ref{definition}, increasing the rank of the concatenated matrix of hidden states among all generated tokens yields a more diverse set of hidden states of the tokens. However, based on \cite{fazel2002matrix}, the optimization of the matrix rank is an NP-hard non-convex problem.  To address this issue, we calculate the nuclear norm of a matrix to approximately measure its rank, as stated in Proposition \ref{proposition2}. Consequently, by denoting the nuclear norm of a matrix as $||\cdot||_{*}$, we can formulate the token diversity loss as follows for verbose images: 
\begin{equation}
\begin{aligned}
\mathcal{L}_{3}(\bm{x}')=-||[g_1(\bm{x}');g_2(\bm{x}');\cdots;g_N(\bm{x}')]||_{*}.
\end{aligned}
\label{eq:diversity loss}
\end{equation}
This token diversity loss can lead to more diverse and complex sequences, making it hard for image-based LLMs to converge to a coherent output.

\begin{proposition}
\label{proposition2}
\cite{fazel2002matrix} \textit{For a rank minimization or maximization problem, the rank of a matrix can be heuristically measured using the nuclear norm of the matrix.}
\end{proposition}

For verbose videos, we propose a frame feature diversity loss tailored for video-based LLMs.
Video-based LLMs extract spatial and temporal features from video data, where different frames maintain a consistent semantic relationship over time. 
The consistent semantic over time in multiple frames within a video induces the sequence generation to converge to  a limited number of tokens. To facilitate a longer generated sequence, we propose to increase the diversity of frame features, thereby introducing inconsistent semantics into the video data. As video-based LLMs grapple with accommodating these diverse or disrupted frame features, it is compelled to explore alternative output interpretations, resulting in a longer generated sequence.

\begin{definition}
\label{definition video}
\textit{Let $\operatorname{Rank}(\cdot)$ indicates the rank of a matrix and $[h_1(\bm{X}');h_2(\bm{X}');\cdots;h_M(\bm{X}')]$ denotes the concatenated matrix of features among all frames. To induce high energy-latency cost of video-based LLMs, the diversity of frame features is defined as the rank of features among all frames, \textit{i.e.}, $\operatorname{Rank}([h_1(\bm{X}');h_2(\bm{X}');\cdots;h_M(\bm{X}')])$.}
\end{definition}
Given by Definition \ref{definition video}, increasing the concatenated matrix of features among all frames can boost the diversity of frame features.  Combined with Proposition \ref{proposition2}, we can formulate the frame feature diversity loss as follows for verbose videos:
\begin{equation}
\begin{aligned}
\mathcal{L}_{3}(\bm{X}')=-||[h_1(\bm{X}');h_2(\bm{X}');\cdots;h_M(\bm{X}')]||_{*},
\end{aligned}
\label{eq:diversity loss videos}
\end{equation}
where $||\cdot||_{*}$ is the nuclear norm of a matrix. This loss can maximize the semantic distance among different frames within a video, which mislead video-based LLMs to generate a longer sequence.
In summary, due to the reduced probability of EOS occurrence by $\mathcal{L}_{1}(\cdot)$, the disruption of the original output dependency introduced by $\mathcal{L}_{2}(\cdot)$, and complexity diversity introduced by $\mathcal{L}_{3}(\cdot)$, our verbose samples can facilitate a more effective manipulation on the worst-case energy-latency cost of MLLMs.

\subsection{Optimization}
\label{sec:loss weight}
To combine the three loss functions of verbose samples, $\mathcal{L}_{1}(\cdot)$, $\mathcal{L}_{2}(\cdot)$, and $\mathcal{L}_{3}(\cdot)$ into an overall objective function, we propose to assign three weights $\lambda_1$, $\lambda_2$, and $\lambda_3$ to the $\mathcal{L}_{1}(\cdot)$, $\mathcal{L}_{2}(\cdot)$, and $\mathcal{L}_{3}(\cdot)$ and sum them up to obtain the overall objective function of verbose images as follows: 
\begin{equation}
\begin{aligned}
\min_{\bm{x}'}\ \lambda_1 \times \mathcal{L}_{1}(\bm{x}') + \lambda_2 \times \mathcal{L}_{2}(\bm{x}') + \lambda_3 \times \mathcal{L}_{3}(\bm{x}'),
\end{aligned}
\label{eq:three losses of final objective}
\end{equation}
and overall objective function of verbose videos as follows:
\begin{equation}
\begin{aligned}
\min_{\bm{X}'}\ \lambda_1 \times \mathcal{L}_{1}(\bm{X}') + \lambda_2 \times \mathcal{L}_{2}(\bm{X}') + \lambda_3 \times \mathcal{L}_{3}(\bm{X}').
\end{aligned}
\label{eq:three losses of final objective videos}
\end{equation}
To optimize this objective, we adopt the projected gradient descent (PGD) algorithm, as proposed by  \cite{madry2017towards}.  
PGD algorithm is an iterative optimization technique that updates the solution by taking steps in the direction of the negative gradient while projecting the result back onto the feasible set. We denote verbose images at the $t$-th step as $\bm{x}'_t$, verbose videos at the $t$-th step as $\bm{X}'_t$, and the gradient descent step of verbose images is as follows:
\begin{equation}
\begin{aligned}
\bm{x}'_{t}=\bm{x}'_{t-1} & -\alpha \times \operatorname{sign} ( \nabla_{\bm{x}'_{t-1}} (\lambda_1  \times \mathcal{L}_{1}(\bm{x}'_{t-1}) \\ &+ \lambda_2 \times \mathcal{L}_{2}(\bm{x}'_{t-1}) + \lambda_3 \times \mathcal{L}_{3}(\bm{x}'_{t-1})) ),\\ & s.t.\ ||\bm{x}'_{t}-\bm{x}||_{p} \le \epsilon,
\end{aligned}
\label{eq:pgd to optimization}
\end{equation}
and the gradient descent step of verbose videos is as follows:
\begin{equation}
\begin{aligned}
\bm{X}'_{t}=\bm{X}'_{t-1} & -\alpha \times \operatorname{sign} ( \nabla_{\bm{X}'_{t-1}} (\lambda_1  \times \mathcal{L}_{1}(\bm{X}'_{t-1}) \\ &+ \lambda_2 \times \mathcal{L}_{2}(\bm{X}'_{t-1}) + \lambda_3 \times \mathcal{L}_{3}(\bm{X}'_{t-1})) ),\\ & s.t.\ \frac{1}{M}\sum_{j=1}^M||\bm{X}_{tj}'-\bm{X}_j||_{p} \le \epsilon,
\end{aligned}
\label{eq:pgd to optimization videos}
\end{equation}
where $\alpha$ is the step size and $\epsilon$ is the perturbation magnitude to ensure the imperceptibility. Since  different loss functions have different convergence rates during the iterative optimization process, we propose a \textbf{\textit{temporal weight adjustment algorithm}} to achieve a better balance among these three loss objectives.  
Specifically, we incorporate normalization scaling and temporal decay functions, $\mathcal{T}_1(t)$, $\mathcal{T}_2(t)$, and $\mathcal{T}_3(t)$, into the optimization weights $\lambda_1(t)$, $\lambda_2(t)$, and $\lambda_3(t)$ of $\mathcal{L}_{1}(\cdot)$, $\mathcal{L}_{2}(\cdot)$, and $\mathcal{L}_{3}(\cdot)$. Therefore, it can be formulated for verbose images as follows: 
\begin{equation}
\begin{aligned}
\lambda_k(t) = ||\mathcal{L}_{2}(\bm{x}'_{t-1})||_1\ /\ ||\mathcal{L}_{k}(\bm{x}'_{t-1})||_1\ /\ \mathcal{T}_k(t), \\
\end{aligned}
\label{eq:dynamic weight of temporal decay}
\end{equation}
and for verbose videos as follows:
\begin{equation}
\begin{aligned}
\lambda_k(t) = ||\mathcal{L}_{2}(\bm{X}'_{t-1})||_1\ /\ ||\mathcal{L}_{k}(\bm{X}'_{t-1})||_1\ /\ \mathcal{T}_k(t), \\
\end{aligned}
\label{eq:dynamic weight of temporal decay videos}
\end{equation}
where $k=1,2,3$ and the temporal decay functions are set as follows:
\begin{equation}
\begin{aligned}
\mathcal{T}_k(t)=a_k \times \operatorname{ln}(t) + b_k.
\end{aligned}
\label{eq:temporal decay function}
\end{equation}
Besides, a momentum value $m$ is introduced into the update process of weights. This involves taking into account not only current weights but also previous weights when updating losses, which helps smooth out the weight updates during the optimization process.

\section{Experiments}

\subsection{Experimental setups}
\textbf{Models and datasets.} For verbose images, we consider four open-source and advanced large image-language models as our evaluation benchmark, including BLIP \cite{li2022blip}, BLIP-2 \cite{li2023blip}, InstructBLIP \cite{dai2023instructblip}, and MiniGPT-4 \cite{zhu2023minigpt}. 
Concretely, we adopt the BLIP with the basic multi-modal mixture of encoder-decoder model in 224M version, BLIP-2 with an OPT-2.7B LM \cite{zhang2022opt}, InstructBLIP and MiniGPT-4 with a Vicuna-7B LM \cite{chiang2023vicuna}. These models perform the captioning task for the image under their prompt templates in default and also evaluate question answering task in discussions.
We randomly choose the 1,000 images from MS-COCO \cite{lin2014microsoft} and ImageNet \cite{deng2009imagenet} dataset, respectively, as our evaluation dataset. 

For verbose videos, we evaluate three benchmark video-based LLMs, including VideoChat-2 \cite{li2023mvbench}, Video-Vicuna \cite{zhang2023video}, and Video-LLaMA \cite{zhang2023video}. Specifically, we employ VideoChat-2 with a Vicuna-7B LM \cite{chiang2023vicuna}, Video-Vicuna with a Vicuna-7B LM \cite{chiang2023vicuna}, and Video-LLaMA a LLaMA-2-7B LM \cite{touvron2023llama2}. These models perform video captioning tasks using their standard prompt templates in default and also evaluate question answering task in discussions. For the evaluation purposes, we randomly select 500 videos from the MSVD \cite{chen2011collecting} and TGIF \cite{jang2017tgif} datasets, respectively.

\textbf{Baselines and setups.}
For the evaluation, we consider original samples, samples with random noise, sponge samples, and NICGSlowDown as baselines. For sponge samples, NICGSlowDown, and our verbose samples, we perform the projected gradient descent (PGD) \cite{madry2017towards} algorithm in $T=1,000$ iterations. 
Besides, in order to ensure the imperceptibility, the perturbation magnitude  is set as 
$\epsilon=8$ within $l_{\infty}$ restriction, following \cite{carlini2019evaluating}, and the step size is set as $\alpha=1$. The default maximum length of generated sequences of MLLMs is set as $512$ and the sampling policy  is configured to use nucleus sampling \cite{holtzman2019curious}. For verbose images, the parameters of loss weights are $a_1=10$, $b_1=-20$, $a_2=0$, $b_2=0$, $a_3=0.5$, and $b_3=1$ and the momentum of our optimization is $m=0.9$. For verbose videos, the parameters of loss weights are $a_1=10,000$, $b_1=100,000$, $a_2=0$, $b_2=0$, $a_3=5$, and $b_3=500$ and the momentum of our optimization is $m=0.9$. Besides, a video consists of eight frames to input the video-based LLMs.



\textbf{Evaluation metrics.} We calculate the energy consumption (J) and the latency time (s) during  inference on one single GPU. Following \cite{shumailov2021sponge}, the energy consumption and latency time are measured by the NVIDIA Management Library (NVML) and the response time cost of an inference, respectively. Besides, the length of generated sequences is also regarded as a metric. Considering the randomness of sampling modes in MLLMs, we report the average evaluation results run over three times.

\begin{table*}[t]
\begin{minipage}{\textwidth}
\caption{The length of generated sequences, energy consumption (J), and latency time (s) of five categories of visual images against four image-based LLMs, including BLIP, BLIP-2, InstructBLIP, and MiniGPT-4, on two datasets, namely MS-COCO and ImageNet. Best results are marked in \textbf{bold}.}
\vspace{-1em}
\label{tab:main results}
\centering
\small
\setlength\tabcolsep{6.75pt}{
\begin{tabular}{@{}ll|ccc|ccc@{}}
\toprule
\multirow{2}{*}{Image-based LLMs} & \multicolumn{1}{l|}{\multirow{2}{*}{Method}} & \multicolumn{3}{c|}{MS-COCO} & \multicolumn{3}{c}{ImageNet} \\ 
 & & Length & Latency & Energy & Length & Latency & Energy  \\
\midrule 
\multirow{5}{*}{BLIP} & Original & 10.03 &  0.21 & 9.51 & 10.17 & 0.22 & 9.10 \\
& Noise & 9.98 & 0.17 & 8.57 & 9.87 & 0.18 & 8.29 \\
& Sponge samples & 65.83 & 1.10 & 73.57 & 76.67 &  1.26 & 86.00 \\
& NICGSlowDown & 179.42 & 2.84 & 220.73 & 193.68 & 2.98 & 243.84 \\
& \textbf{Verbose images (Ours)} &  \textbf{318.66} & \textbf{5.13} & \textbf{406.65} & \textbf{268.25} & \textbf{4.31} & \textbf{344.91} \\
\midrule
\multirow{5}{*}{BLIP-2} & Original & 8.82 & 0.39 & 16.08 & 8.11 & 0.37 & 15.39 \\
& Noise & 9.55 & 0.43 & 17.53 & 8.37 & 0.44 & 19.39 \\
& Sponge samples & 22.53 & 0.73 & 30.20 & 43.59 & 1.51 & 63.27 \\
& NICGSlowDown & 103.54 & 3.78 & 156.61 & 129.68 & 4.34 & 180.06 \\
& \textbf{Verbose images (Ours)} & \textbf{226.72} & \textbf{7.97} & \textbf{321.59} & \textbf{250.72} & \textbf{10.26} & \textbf{398.58} \\
\midrule
\multirow{5}{*}{InstructBLIP} & Original & 63.79 & 2.97 & 151.80 & 54.40 & 2.60 & 128.03 \\
& Noise & 62.76 & 2.91 & 148.64 &  53.01 & 2.50 & 125.42 \\
& Sponge samples & 92.69 & 4.10 & 209.81 & 80.26 & 3.55 & 175.17 \\
& NICGSlowDown & 93.70 & 4.08 &  200.51 & 81.64 & 3.56 & 174.44 \\
& \textbf{Verbose images (Ours)} & \textbf{140.35} & \textbf{6.15} & \textbf{316.06} & \textbf{131.79}  & \textbf{6.05} & \textbf{300.43} \\
\midrule
\multirow{5}{*}{MiniGPT-4} & Original &  45.29 & 10.39 &  329.50 & 40.93 &  9.11 & 294.68   \\
& Noise & 45.15 & 10.35 & 327.04 & 47.78 & 10.98 & 348.66 \\
& Sponge samples & 220.30 & 43.84 & 1390.73 & 228.70 & 47.74 & 1528.58 \\
& NICGSlowDown & 232.80 & 46.39 & 1478.74 & 245.51 &  51.22 & 1624.06 \\
& \textbf{Verbose images (Ours)} & \textbf{321.35} & \textbf{67.14} & \textbf{2113.29} & \textbf{321.24} & \textbf{64.31} &  \textbf{2024.62} \\
\bottomrule
\end{tabular}}
\end{minipage}
\begin{minipage}{\textwidth}
\caption{The length of generated sequences, energy consumption (J), and latency time (s) of five categories of visual videos against three video-based LLMs, including VideoChat-2, Video-Vicuna, and Video-LLaMA, on two datasets, namely MSVD and TGIF. Best results are marked in \textbf{bold}.}
\vspace{-1em}
\label{tab:main results videos}
\centering
\small
\setlength\tabcolsep{6.9pt}{
\begin{tabular}{@{}ll|ccc|ccc@{}}
\toprule
\multirow{2}{*}{Video-based LLMs} & \multicolumn{1}{l|}{\multirow{2}{*}{Method}} & \multicolumn{3}{c|}{MSVD} & \multicolumn{3}{c}{TGIF} \\ 
 & & Length & Latency & Energy & Length & Latency & Energy  \\
\midrule 
\multirow{5}{*}{VideoChat-2} & Original & 13.97	& 4.79 & 153.08 & 14.47& 5.16 & 163.22 \\
& Noise & 14.49 & 4.84 & 156.84 & 14.38 & 4.76 & 148.98 \\
& Sponge samples & 28.84 & 6.99 & 222.83 & 25.02 & 6.46 & 202.93 \\
& NICGSlowDown & 29.99 & 7.02 & 224.11 & 24.98 & 6.35 & 197.22 \\
& \textbf{Verbose videos (Ours)} & \textbf{165.77} & \textbf{35.88} & \textbf{1323.42} & \textbf{198.43} & \textbf{46.07} & \textbf{1741.97} \\
\midrule
\multirow{5}{*}{Video-Vicuna} & Original & 77.18 & 13.05 & 429.36 & 83.67 & 13.88 & 441.52  \\
& Noise & 80.21 & 13.45 & 444.16 & 85.82 & 14.51 & 472.84 \\
& Sponge samples & 225.91 & 39.05 & 1285.56 & 231.25 & 39.37 & 1315.16 \\
& NICGSlowDown & 230.63 &  41.03 & 1310.19 & 235.22 & 42.59 & 1380.44 \\
& \textbf{Verbose videos (Ours)} & \textbf{262.97} & \textbf{45.52} & \textbf{1488.95} & \textbf{277.56} & \textbf{47.88} & \textbf{1513.22} \\
\midrule
\multirow{5}{*}{Video-LLaMA} & Original & 59.51 & 12.01 & 441.58 & 60.07 & 13.02 & 495.73 \\
& Noise & 63.52 & 14.03 & 578.78 & 61.82 & 13.62 & 525.28 \\
& Sponge samples & 145.92 & 30.08 & 1193.59 & 136.91 & 28.69 & 1061.85 \\
& NICGSlowDown & 172.92 & 37.56 & 1408.79 & 171.28 & 38.36 & 1616.52 \\
& \textbf{Verbose videos (Ours)} & \textbf{180.01} & \textbf{41.15} & \textbf{1815.22} & \textbf{179.24} & \textbf{39.83} & \textbf{1726.57} \\
\bottomrule
\end{tabular}}
\end{minipage}
\end{table*}


\begin{figure*}[t]
\begin{minipage}{\textwidth}
\centering
\subfloat[BLIP] {
\label{BLIP of length distribution}  
\includegraphics[width=0.232\textwidth]{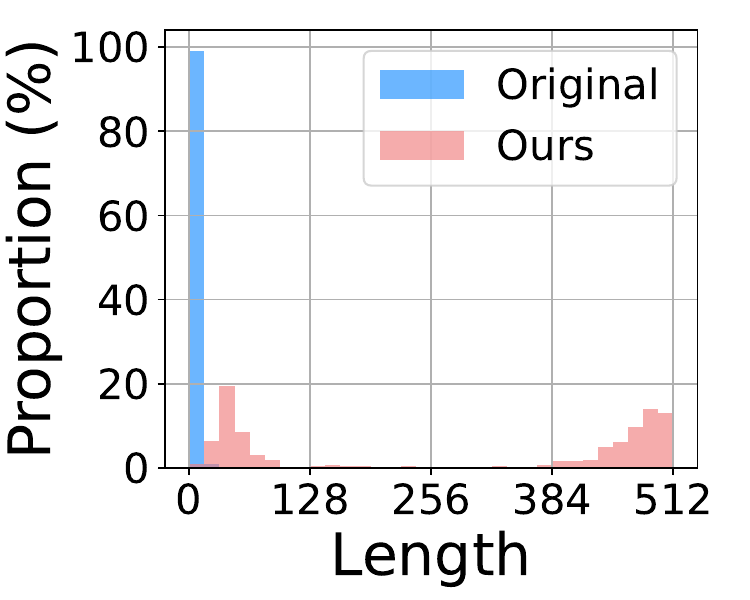}   
}        
\subfloat[BLIP-2] {   
\label{BLIP2 of length distribution}  \includegraphics[width=0.232\textwidth]{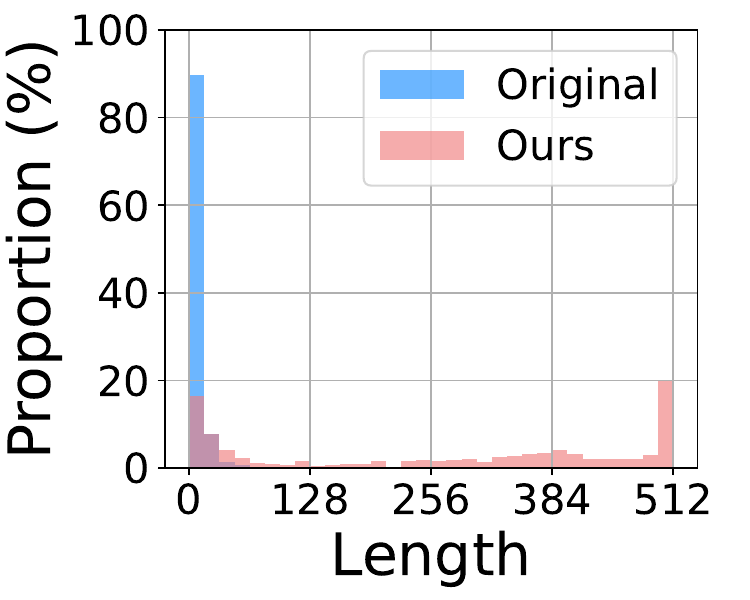}   
}    
\subfloat[InstructBLIP] {   
\label{InstructBLIP of length distribution}  \includegraphics[width=0.232\textwidth]{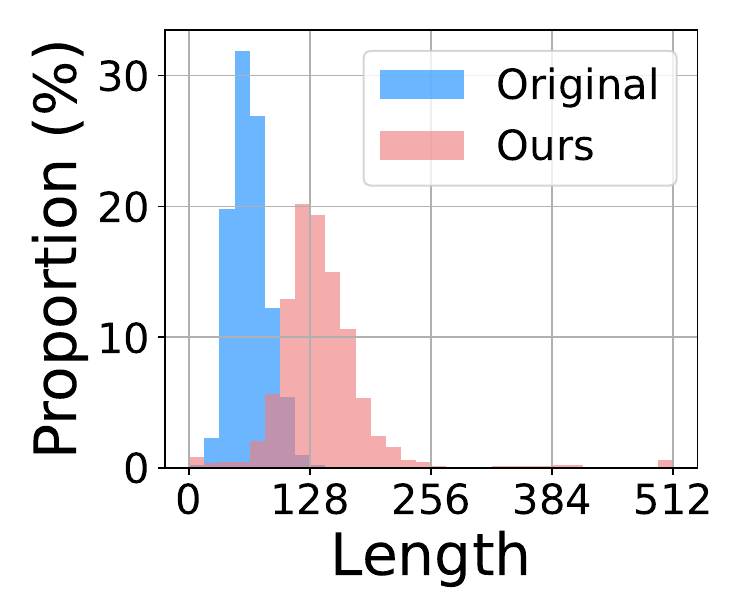}  
}    
\subfloat[MiniGPT-4] {   
\label{MiniGPT-4 of length distribution}  \includegraphics[width=0.232\textwidth]{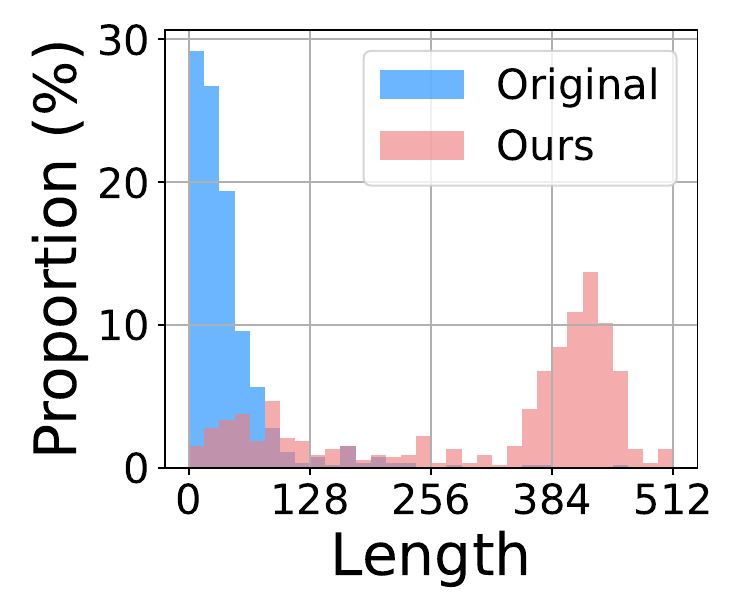}  
}    
\caption{The length distribution of four image-based LLMs: (a) BLIP. (b) BLIP-2. (c) InstructBLIP. (d) MiniGPT-4. The peak of length distribution of our verbose images shifts towards longer sequences.} 
\label{all length distribution}
\end{minipage}
\vspace{-1em}
\begin{minipage}{\textwidth}
\centering
\subfloat[VideoChat-2] {
\label{VideoChat-2 of length distribution}  
\includegraphics[width=0.232\textwidth]{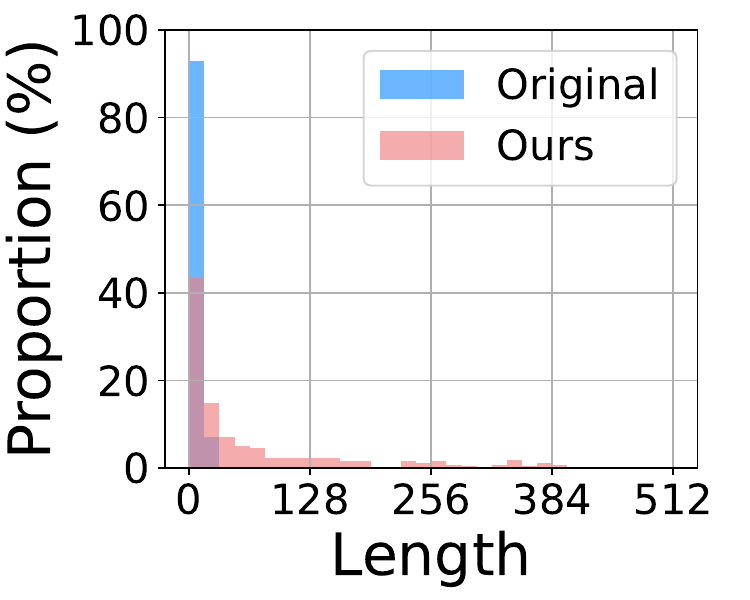}   
}        
\subfloat[Video-Vicuna] {   
\label{Video-Vicuna of length distribution}  \includegraphics[width=0.232\textwidth]{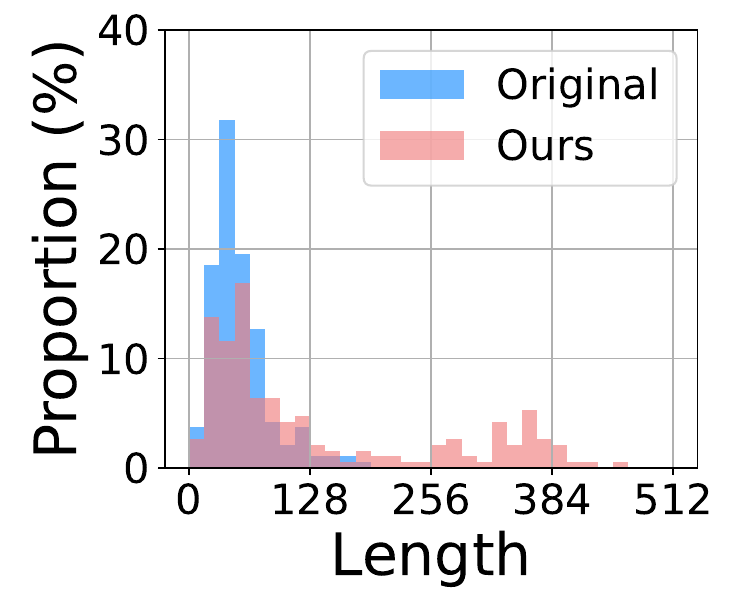}   
}    
\subfloat[Video-LLaMA] {   
\label{Video-LLaMA of length distribution}  \includegraphics[width=0.232\textwidth]{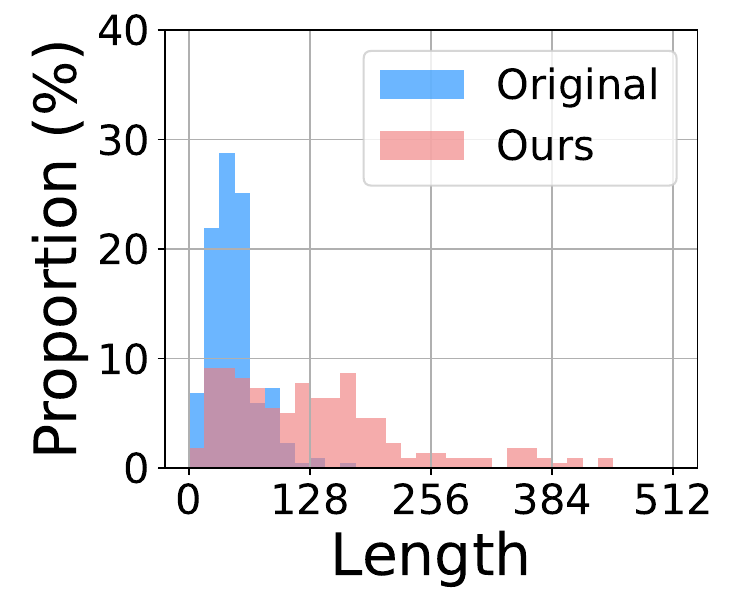}  
}    
\caption{The length distribution of three video-based LLMs: (a) VideoChat-2. (b) Video-Vicuna. (c) Video-LLaMA. The peak of length distribution of our verbose videos shifts towards longer sequences.} 
\label{all length distribution videos}
\end{minipage}
\vspace{-1em}
\end{figure*}

\begin{table*}[t]
\begin{minipage}{\textwidth}
\caption{The length of generated sequences, energy consumption (J), and latency time (s) against BLIP-2 of verbose images on MS-COCO and ImageNet, and against VideoChat-2 of verbose videos on MSVD and TGIF in different combinations of three loss objectives. The total of three losses can induce MLLMs to generate the longest sequences. }
\vspace{-1em}
\label{tab:ablation for loss functions}
\centering
\small
\setlength\tabcolsep{4.5pt}{
\begin{tabular}{@{}ccc|ccc|ccc|ccc|ccc@{}}
\toprule
\multirow{2}{*}{$\mathcal{L}_{1}$} & \multirow{2}{*}{$\mathcal{L}_{2}$} & \multicolumn{1}{c|}{\multirow{2}{*}{$\mathcal{L}_{3}$}} & \multicolumn{3}{c|}{MS-COCO} & \multicolumn{3}{c|}{ImageNet} & \multicolumn{3}{c|}{MSVD} & \multicolumn{3}{c}{TGIF} \\ 
&  & & Length & Latency & Energy & Length & Latency & Energy & Length & Latency & Energy & Length & Latency & Energy \\
\midrule 
\checkmark &  &  & 119.46 &  3.96 & 162.40 & 147.87 & 4.52 & 185.64 & 27.66 & 6.84 & 227.12 & 24.87 & 6.41 & 203.18 \\
& \checkmark &  & 139.54 & 4.65 & 194.17 & 161.46 & 5.69 & 240.25 & 119.94 & 23.68 & 773.09 & 132.84 & 25.48 & 848.63 \\
&  & \checkmark & 104.03 &  3.29 & 135.75 & 129.02 & 3.90 & 161.87 & 91.97 & 18.62 & 588.26 & 111.13 & 23.35 & 765.24 \\
\checkmark & \checkmark &  & 177.95 & 6.47 & 267.01 & 217.78 & 7.47 & 306.09 & 127.02 & 25.64 & 859.44 & 165.25 & 36.16 & 1244.91 \\
\checkmark &  & \checkmark & 150.79 & 4.51 & 182.16 & 151.57 & 4.71 & 194.40 & 102.17 & 20.28 & 695.03 & 141.74 & 32.84 & 1148.93 \\
& \checkmark & \checkmark & 176.53 & 6.05 & 254.30 & 206.43 & 7.50 & 304.06  & 133.17 & 25.22 & 892.05 & 184.21 & 39.09 & 1316.51 \\
\checkmark & \checkmark & \checkmark & \textbf{226.72} & \textbf{7.97} & \textbf{321.59} & \textbf{250.72} & \textbf{10.26} & \textbf{398.58} & \textbf{165.77} & \textbf{35.88} & \textbf{1323.42} & \textbf{198.43} & \textbf{46.07} & \textbf{1741.97} \\
\bottomrule
\end{tabular}}
\end{minipage}
\begin{minipage}{\textwidth}
\caption{The length of generated sequences, energy consumption (J), and latency time (s) against BLIP-2 of verbose images on MS-COCO and ImageNet, and against VideoChat-2 of verbose videos on MSVD and TGIF in different combinations of two optimization modules. The total of two modules can induce MLLMs to generate the longest sequences. }
\vspace{-1em}
\label{tab:ablation for the optimization}
\centering
\small
\setlength\tabcolsep{5.1pt}{
\begin{tabular}{@{}cc|ccc|ccc|ccc|ccc@{}}
\toprule
\multirow{2}{*}{$\mathcal{T}(t)$} & \multicolumn{1}{c|}{\multirow{2}{*}{$m$}} & \multicolumn{3}{c|}{MS-COCO} & \multicolumn{3}{c|}{ImageNet} & \multicolumn{3}{c|}{MSVD} & \multicolumn{3}{c}{TGIF} \\  
& & Length & Latency & Energy & Length & Latency & Energy & Length & Latency & Energy & Length & Latency & Energy  \\
\midrule 
 &  &  152.49 & 4.70 & 205.09 & 144.90 & 5.31 & 231.83 &  128.55 & 24.25 & 853.35 & 151.83 & 35.51 & 1282.74 \\
\checkmark &  & 199.92 & 7.02 & 292.55 & 231.03 & 7.88 & 318.34 & 144.51 & 33.19 & 1163.06 & 187.69 & 39.29 & 1390.18 \\
 & \checkmark & 187.32 & 6.89 & 274.67 & 214.92 & 7.49 & 308.11 & 130.22 & 25.83 & 911.61 & 162.51 & 34.74 & 1204.18  \\
\checkmark & \checkmark  & \textbf{226.72} & \textbf{7.97} & \textbf{321.59} & \textbf{250.72} & \textbf{10.26} & \textbf{398.58} & \textbf{165.77} & \textbf{35.88} & \textbf{1323.42} & \textbf{198.43} & \textbf{46.07} & \textbf{1741.97} \\
\bottomrule
\end{tabular}}
\end{minipage}
\vspace{-0.5em}
\end{table*}

\subsection{Main results}
Table \ref{tab:main results} and Table \ref{tab:main results videos} compare the length of generated sequences, energy consumption, and latency time of original samples, samples with random noise, sponge samples, NICGSlowdown, and our proposed verbose samples. 
The original samples serve as a baseline, providing reference values for comparison. 
When random noise is added to the samples, the generated sequences exhibit a similar length to those of the original ones. It illustrates that it is necessary to optimize a handcrafted perturbation to induce high energy-latency cost of MLLMs. The sponge samples and NICGSlowdown can generate longer sequences compared to original ones. However, the increase in length is still smaller than that of our verbose samples. This can be attributed to the reason that the additional computational cost brought by sponge samples is tailored for LLMs and the objective for longer sequences introduced by NICGSlowdown is particularly suited for smaller-scale captioning models. There is still potential improvement to induce high energy-latency cost specifically for MLLMs.

Our verbose samples can increase the maximum length of generated sequences and introduce the highest energy-latency cost among all these methods.  Specifically, our verbose images can increase the average length of generated sequences by 7.87$\times$ and 8.56$\times$ relative to original images on the MS-COCO and ImageNet datasets, respectively. Our verbose videos can increase the average length of generated sequences by 4.04$\times$ and 4.14$\times$ compared to original videos on MSVD and TGIF datasets. These results demonstrate the superiority of our verbose samples for MLLMs.
In addition, we visualize the length distribution of output sequences generated by four image-based LLMs on original images and our verbose images in Fig. \ref{all length distribution} and that by three video-based LLMs on original videos and our verbose videos in Fig. \ref{all length distribution videos}. Compared to original samples, the distribution peak for sequences generated using our verbose samples exhibits a shift towards the direction of the longer length, confirming the effectiveness of our methods in generating longer sequences. We conjecture that the different shift magnitudes are due to different architectures, different training policies, and different parameter quantities in these MLLMs. 

\begin{table*}
\begin{minipage}{\textwidth}
\caption{The length of generated sequences, energy consumption (J), and latency time (s) of different attacking methods, which compares differences between verbose images and videos when inducing high energy-latency cost of video-based LLMs.}
\vspace{-1em}
\label{tab:difference between verbose images and videos}
\centering
\small
\setlength\tabcolsep{11pt}{
\begin{tabular}{@{}l|ccc|ccc@{}}
\toprule
\multirow{2}{*}{Attacking method} & \multicolumn{3}{c|}{MSVD} & \multicolumn{3}{c}{TGIF} \\
 & Length & Latency & Energy & Length & Latency & Energy  \\
\midrule 
$L_{3\_img}(\cdot)$ & 48.20 & 10.16 & 329.61 & 58.45 &  12.04 & 407.86 \\
$L_{3\_vid}(\cdot)$ & 91.97 & 18.62 & 588.26 & 111.13 & 23.35 & 765.24 \\
Verbose videos using $L_{3\_img}(\cdot)$ & 133.93 & 27.99 &  1194.35 & 176.15 & 39.72 &  1582.74 \\
\midrule 
Verbose images & 71.35 & 14.08 & 448.07 & 63.73 & 12.52 & 407.95 \\
Multiple verbose images & 104.72 & 21.52 & 638.28 & 127.37 & 27.22 & 852.51 \\
\textbf{Verbose videos (Ours)} & \textbf{165.77} & \textbf{35.88} & \textbf{1323.42} & \textbf{198.43} & \textbf{46.07} & \textbf{1741.97} \\
\bottomrule
\end{tabular}
}
\end{minipage}
\vspace{-1em}
\end{table*}

\begin{table*}
\begin{minipage}{\textwidth}
\caption{The $\text{CHAIR}_i$ (\%) and $\text{CHAIR}_s$ (\%) of the original images and our verbose images against four image-based LLMs. Our verbose images achieve a higher both $\text{CHAIR}_i$ and $\text{CHAIR}_s$, meaning more hallucinated objects.}
\vspace{-1em}
\label{tab:object hallucination}
\centering
\small
\setlength\tabcolsep{9.2pt}{
\begin{tabular}{@{}l|cccc|cccc@{}}
\toprule
\multirow{3}{*}{Image-based LLMs} & \multicolumn{4}{c|}{$\text{CHAIR}_i$ (\%)} & \multicolumn{4}{c}{$\text{CHAIR}_s$ (\%)} \\
& \multicolumn{2}{c}{MS-COCO} & \multicolumn{2}{c|}{ImageNet} & \multicolumn{2}{c}{MS-COCO} & \multicolumn{2}{c}{ImageNet}  \\ 
& Original & Ours  & Original & Ours & Original & Ours  & Original & Ours \\
\midrule 
BLIP & 11.41 & 79.93 & 22.29 & 89.80 & 12.77 & 84.22 & 13.77 & 90.33 \\
BLIP-2 & 12.03 & 52.30 & 25.30 & 69.83 & 10.99 & 35.02 & 11.77 & 46.11 \\
InstructBLIP & 23.66 & 55.56 & 40.11 & 69.27 & 38.04 & 75.46 & 34.55 & 64.55 \\
MiniGPT-4 & 19.42 & 46.65 & 29.20 & 65.50 & 19.61 & 52.01 & 16.57 & 54.37 \\
\bottomrule
\end{tabular}}
\end{minipage}
\begin{minipage}{\textwidth}
\caption{The $\text{CHAIR}_i$ (\%) and $\text{CHAIR}_s$ (\%) of the original videos and our verbose videos against three video-based LLMs. Our verbose videos achieve a higher both $\text{CHAIR}_i$ and $\text{CHAIR}_s$, meaning more hallucinated objects.}
\vspace{-1em}
\label{tab:object hallucination of videos}
\centering
\small
\setlength\tabcolsep{9.2pt}{
\begin{tabular}{@{}l|cccc|cccc@{}}
\toprule
\multirow{3}{*}{Video-based LLMs} & \multicolumn{4}{c|}{$\text{CHAIR}_i$ (\%)} & \multicolumn{4}{c}{$\text{CHAIR}_s$ (\%)} \\
& \multicolumn{2}{c}{MSVD} & \multicolumn{2}{c|}{TGIF} & \multicolumn{2}{c}{MSVD} & \multicolumn{2}{c}{TGIF}  \\ 
& Original & Ours  & Original & Ours & Original & Ours  & Original & Ours \\
\midrule 
VideoChat-2 & 27.53 & 37.37 & 19.81 & 37.22 & 6.58 & 12.4 & 7.91 & 15.41 \\
Video-Vicuna & 40.33 & 53.64 & 42.51 & 47.96 & 55.45 & 63.54 & 48.52 & 56.07  \\
Video-LLaMA & 45.86 & 64.43 & 51.67 & 58.26 & 55.25 & 74.42 & 52.57 & 59.89  \\
\bottomrule
\end{tabular}}
\end{minipage}
\end{table*}

\subsection{Ablation studies}
We explore the effect of the proposed three loss objectives and  the effect of the temporal weight adjustment algorithm with momentum for verbose samples.

\textbf{Effect of loss objectives.} Our verbose samples consist of three loss objectives: $\mathcal{L}_{1}(\cdot)$, $\mathcal{L}_{2}(\cdot)$ and $\mathcal{L}_{3}(\cdot)$, respectively. Note that $\mathcal{L}_{3}(\cdot)$ is different and modality specific for verbose images and videos. To identify the individual contributions of each loss function and their combined effects on the overall performance, we evaluate various combinations of the proposed loss functions, as presented in Table \ref{tab:ablation for loss functions}. It can be observed that optimizing each loss function individually can generate longer sequences, and the combination of all three loss functions achieves the best results in terms of sequence length. This ablation study suggests that the three loss functions in verbose images, which delay EOS occurrence, enhance output uncertainty, and improve token diversity, play a complementary role in extending the length of generated sequences. Similarly, the three loss functions in verbose videos, which delay EOS occurrence, enhance output uncertainty, and improve frame feature diversity, collaboratively contribute to extending the length of generated sequences.

\textbf{Effect of temporal weight adjustment.} During the optimization, we introduce two methods: a temporal decay $\mathcal{T}(t)$ for loss weighting and an addition of the momentum $m$. As shown in Table \ref{tab:ablation for the optimization}, both methods contribute to the longer length of the generated sequences. Furthermore, the longest length is obtained by combining temporal decay and momentum, which indicates that temporal decay and momentum can work synergistically to induce high energy-latency cost for MLLMs.

\subsection{Differences between verbose images and videos}
To induce high energy-latency cost of video-based LLMs, we do not directly use the previously proposed verbose images. Instead, we introduce verbose videos, which are specifically tailored for video-based LLMs. To illustrate the reason behind it, we design two experiments as follows to compare differences between verbose images and videos.

(1) From the difference of loss objectives, verbose images and videos have a different modality specific $L_3(\cdot)$. For clearer expression, $L_3(\cdot)$ in verbose images and videos are abbreviated as $L_{3\_img}(\cdot)$ and $L_{3\_vid}(\cdot)$, respectively. Therein, $L_{3\_vid}(\cdot)$ aims to destroy the diversity of frame features, targeting specific video data with multiple frames rather than image data, making it inapplicable to image-based LLMs. Conversely, $L_{3\_img}(\cdot)$ in verbose images aims to improve the diversity of hidden states among all generated tokens, which can be applied to both image-based LLMs and video-based LLMs because they both employ auto-regressive token generation. Therefore, we explore the effect when replacing $L_{3\_vid}(\cdot)$ with $L_{3\_img}(\cdot)$ to generate verbose videos. Results in Table \ref{tab:difference between verbose images and videos} reveal that using $L_{3\_vid}(\cdot)$ alone outperforms solely employing $L_{3\_img}(\cdot)$. Correspondingly, verbose videos with $L_{3\_vid}(\cdot)$ exhibit a superior length of generated sequences, indicating its effectiveness in generating better verbose videos compared with $L_{3\_img}(\cdot)$.

(2) Considering various ways to utilize verbose images, we investigate the impact of using either a single verbose image or a combination of multiple verbose images to form a video, compared to the direct application of verbose videos.  
Specifically, we conduct an experiment where we randomly select a frame from a video and use it to generate a verbose image. Additionally, we randomly select eight frames, generate a corresponding verbose image for each, and then combine these to form a video, the same frame number as our verbose videos. 
As demonstrated in Table \ref{tab:difference between verbose images and videos}, the generated sequences of both methods are lower compared to our verbose videos. We conjecture the reason is that a single verbose image or multiple verbose images do not take into account the temporal relationship among different frames within a video. 
Our verbose videos, on the other hand, consider it using $L_{3\_vid}(\cdot)$ and introduce inconsistent semantics, which is a crucial factor in the effectiveness of the energy-latency manipulation for video-based LLMs.

\begin{figure*}[t] \centering   
\begin{minipage}{\textwidth}
\includegraphics[width=\columnwidth]{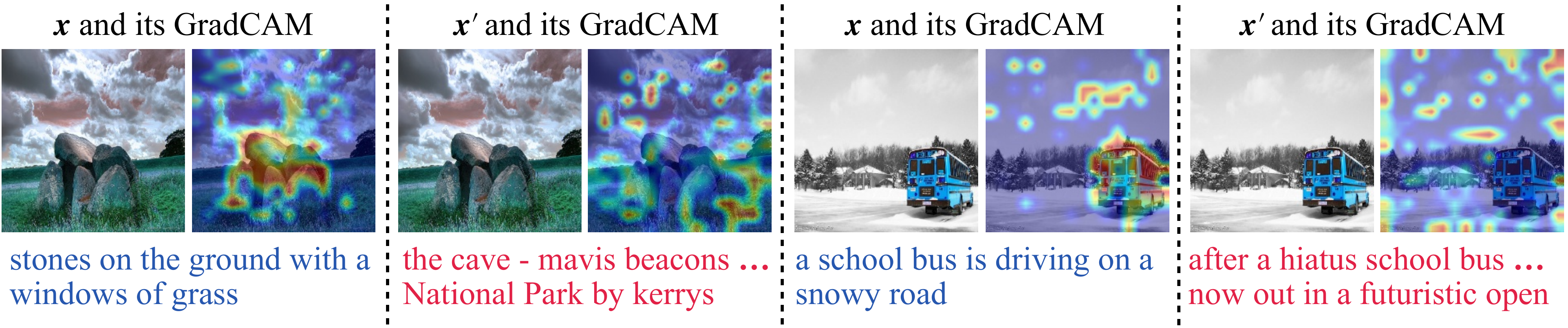}  
\vspace{-2em}
\caption{GradCAM for the original image $\bm{x}$ and our verbose counterpart $\bm{x}'$. The attention of our verbose images is more dispersed and uniform. } 
\label{grad cam image}
\end{minipage}
\vspace{2em}
\begin{minipage}{\textwidth}
\includegraphics[width=\columnwidth]{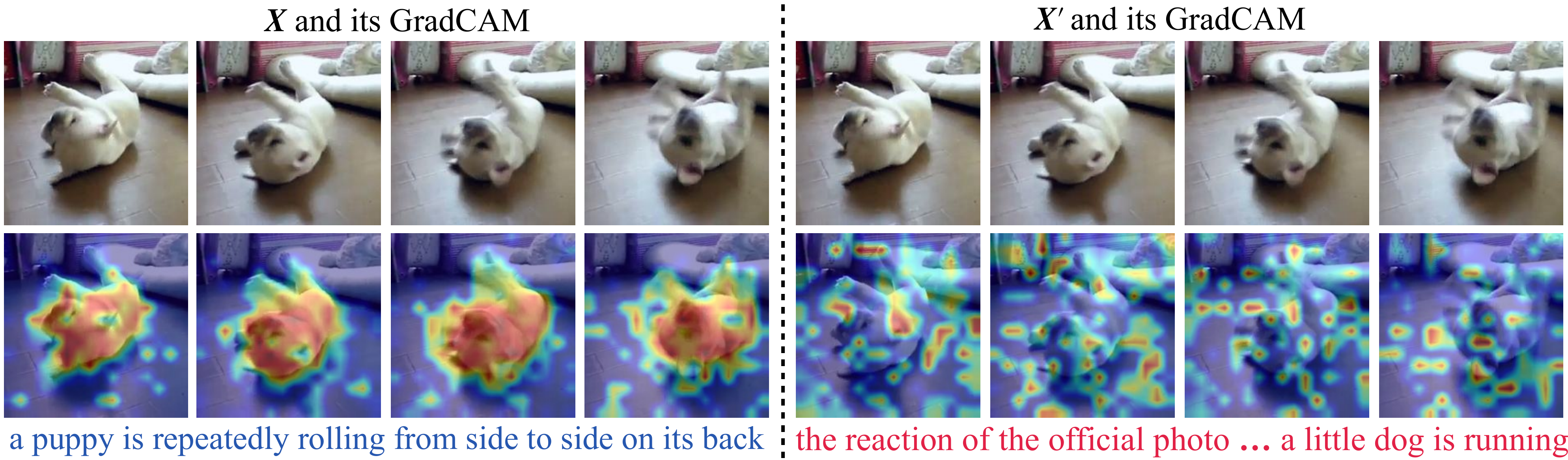} 
\vspace{-2em}
\caption{GradCAM for the original video $\bm{X}$ and our verbose counterpart $\bm{X}'$. The attention of our verbose videos is more dispersed and uniform.  } 
\label{grad cam video}
\end{minipage}
\vspace{-3.5em}
\label{grad cam}
\end{figure*}

\begin{table*}
\begin{minipage}{\textwidth}
\caption{The length of generated sequences, energy consumption (J), latency time (s), and LIPIS against BLIP-2 on MS-COCO and ImageNet of verbose images and against VideoChat-2 on MSVD and TGIF of verbose videos with different perturbation magnitudes $\epsilon$. A larger magnitude leads to a longer sequence and a more perceptible image.}
\vspace{-1em}
\label{tab:varying perturbation magnitude}
\centering
\small
\setlength\tabcolsep{2pt}{
\begin{tabular}{@{}c|cccc|cccc|cccc|cccc@{}}
\toprule
\multirow{2}{*}{$\epsilon$} & \multicolumn{4}{c|}{MS-COCO} & \multicolumn{4}{c|}{ImageNet} & \multicolumn{4}{c|}{MSVD} & \multicolumn{4}{c}{TGIF} \\ 
& Length & Latency & Energy & LIPIS & Length & Latency & Energy & LIPIS & Length & Latency & Energy & LIPIS & Length & Latency & Energy & LIPIS  \\
\midrule 
2 & 91.75 & 3.06 & 126.22 & 0.003 & 103.51 & 3.50 & 144.30 & 0.003 & 27.28 & 7.12 & 242.45 & 0.030 & 26.15 & 6.82 & 225.98 & 0.036 \\
4 & 141.46 & 4.63 & 187.14 & 0.012 & 147.30 & 4.90 & 199.24 & 0.013 & 41.62 & 11.56 & 414.59 & 0.104 & 41.82 & 10.51 & 400.14 & 0.116 \\
8 & 226.72 & 7.97 & 321.59 & 0.036 & 250.72 & 10.26 & 398.58 & 0.037 & 165.77 & 35.88 & 1323.42 & 0.382 & 198.43 & 46.07 & 1741.97 & 0.366 \\
16 & 251.09 & 8.41 & 355.00 & 0.087 & 272.95 & 9.51 & 380.86 & 0.086 & 450.14 & 96.34 & 3541.62 & 0.802 & 478.18 & 102.40 & 3865.42 & 0.827 \\
32 & 287.22 & 9.13 & 377.64 & 0.160 & 321.65 & 10.61 &  429.77 & 0.157 & 508.21 & 110.60 & 3742.91 & 1.600 & 510.05 & 115.21 & 4089.93 & 1.701 \\
\bottomrule
\end{tabular}}
\end{minipage}
\end{table*}

\begin{table*}[t]
\caption{The length of generated sequences, energy consumption (J), and latency time (s) of black-box  transferability across four image-based LLMs of our verbose images. Our verbose images can transfer across different image-based LLMs. }
\vspace{-1em}
\label{tab:transferability of four models of our verbose images}
\centering
\small
\setlength\tabcolsep{9.5pt}{
\begin{tabular}{@{}ll|ccc|ccc@{}}
\toprule
\multirow{2}{*}{Source model} & \multicolumn{1}{l|}{\multirow{2}{*}{Target model}} & \multicolumn{3}{c|}{MS-COCO} & \multicolumn{3}{c}{ImageNet} \\
 & & Length & Latency & Energy & Length & Latency & Energy  \\
\midrule 
None & \multirow{5}{*}{BLIP} & 10.03 &  0.21 & 9.51 & 10.17 & 0.22 & 9.10 \\
BLIP &  &  \textbf{318.66} & \textbf{5.13} & \textbf{406.65} & \textbf{268.25} & \textbf{4.31} & \textbf{344.91} \\
BLIP-2 &  & 14.51 & 0.24 & 10.05 & 14.03 & 0.24 & 10.23 \\
InstructBLIP &  &  63.43 & 2.84 & 142.46 & 54.14 & 2.52 & 131.22 \\
MiniGPT-4 &  & 48.50 & 10.23 & 316.28 & 49.14 & 10.20 & 321.29 \\
\midrule 

None & \multirow{5}{*}{BLIP-2} & 8.82 & 0.39 & 16.08 & 8.11 & 0.37 & 15.39 \\
BLIP &  & 36.09 & 1.19 &  47.07 & 73.22 & 2.39 & 99.24 \\
BLIP-2 &  & \textbf{226.72} & \textbf{7.97} & \textbf{321.59} & \textbf{250.72} & \textbf{10.26} & \textbf{398.58} \\
InstructBLIP &  & 140.05 & 3.91 & 166.40 & 145.39 & 4.07 & 175.01 \\
MiniGPT-4 &  & 140.88 & 3.81 & 154.43 & 140.92 & 3.91 & 165.95 \\
\midrule 

None & \multirow{5}{*}{InstructBLIP} & 63.79 & 2.97 & 151.80 & 54.40 & 2.60 & 128.03 \\
BLIP &   & 91.94 & 4.13 & 203.94 & 82.66 & 3.77 & 186.51 \\
BLIP-2 &  & 109.01 & 4.87 & 240.30 & 99.25 & 4.53 & 225.55 \\
InstructBLIP &  & \textbf{140.35} & \textbf{6.15} & \textbf{316.06} & \textbf{131.79}  & \textbf{6.05} & \textbf{300.43} \\
MiniGPT-4 &  & 100.08 & 4.42 & 210.58 & 99.42 & 4.47 & 219.08 \\
\midrule 

None & \multirow{5}{*}{MiniGPT-4} & 45.29 & 10.39 &  329.50 & 40.93 &  9.11 & 294.68 \\
BLIP &  & 229.10 & 48.90 & 1562.25 & 254.57 & 54.57 & 1691.51\\
BLIP-2 &  & 296.77 & 58.84 & 1821.66 & 289.19 & 58.79 & 1826.81 \\
InstructBLIP &  & 270.73 & 48.88 & 1551.04 & 258.32 & 50.26 & 1632.01 \\
MiniGPT-4 &   & \textbf{321.35} & \textbf{67.14} & \textbf{2113.29} & \textbf{321.24} & \textbf{64.31} &  \textbf{2024.62} \\
\bottomrule
\end{tabular}}
\vspace{-1em}
\end{table*}

\begin{table*}[t]
\caption{The length of generated sequences, energy consumption (J), and latency time (s) of black-box  transferability across three video-based LLMs of our verbose videos. Our verbose videos can transfer across different video-based LLMs. }
\vspace{-1em}
\label{tab:transferability of four models of our verbose images of videos}
\centering
\small
\setlength\tabcolsep{9pt}{
\begin{tabular}{@{}ll|ccc|ccc@{}}
\toprule
\multirow{2}{*}{Source model} & \multicolumn{1}{l|}{\multirow{2}{*}{Target model}} & \multicolumn{3}{c|}{MSVD} & \multicolumn{3}{c}{TGIF} \\
 & & Length & Latency & Energy & Length & Latency & Energy  \\
\midrule 
None & \multirow{4}{*}{VideoChat-2} & 13.97	& 4.79 & 153.08 & 14.47& 5.16 & 163.22 \\
VideoChat-2 &  &  \textbf{165.77} & \textbf{35.88} & \textbf{1323.42} & \textbf{198.43} & \textbf{46.07} & \textbf{1741.97}   \\
Video-Vicuna &  & 72.53 & 13.48 & 412.85 & 78.28 & 15.03 & 433.29 \\
Video-LLaMA &  & 63.15 & 11.43 & 349.65 & 61.31 & 12.14 & 353.93 \\
\midrule 

None & \multirow{4}{*}{Video-Vicuna} & 77.18 & 13.05 & 429.36 & 83.67 & 13.88 & 441.52  \\
VideoChat-2 &  & 90.65 & 15.26 & 503.73 & 105.23 & 16.46 & 532.66 \\
Video-Vicuna &  & \textbf{262.97} & \textbf{45.52} & \textbf{1488.95} & \textbf{277.56} & \textbf{47.88} & \textbf{1513.22} \\
Video-LLaMA &  &  102.46 & 17.32 & 539.88 & 125.83 & 18.85 & 563.25 \\
\midrule 

None & \multirow{4}{*}{Video-LLaMA} & 59.51 & 12.01 & 441.58 & 60.07 & 13.02 & 465.73  \\
VideoChat-2 &  & 84.63 & 14.91 & 462.36 & 87.03 & 15.48 & 494.05 \\
Video-Vicuna &  &  102.74 & 17.53 & 536.52 & 110.32 & 18.86 & 548.85 \\
Video-LLaMA &  & \textbf{180.01} & \textbf{41.15} & \textbf{1815.22} & \textbf{179.24} & \textbf{39.83} & \textbf{1726.57} \\

\bottomrule
\end{tabular}}
\end{table*}

\subsection{Unified interpretation}
To better reveal the mechanisms behind our verbose samples, we propose a unified interpretation, which conducts two further studies, including the visual interpretation where we adopt Grad-CAM \cite{selvaraju2017grad} to generate the attention maps and the textual interpretation where we evaluate the object hallucination in generated sequences by CHAIR \cite{rohrbach2018object}. 

\textbf{Textual Interpretation.} 
We investigate object hallucination in generated sequences using CHAIR \cite{rohrbach2018object}. 
$\text{CHAIR}_i$ is calculated as the fraction of hallucinated object instances, while $\text{CHAIR}_s$ represents the fraction of sentences containing a hallucinated object, with the results presented in Table \ref{tab:object hallucination} and Table \ref{tab:object hallucination of videos}.
Compared to original samples, which exhibit a lower object hallucination rate, the longer sequences produced by our verbose samples contain a broader set of objects. This observation implies that our verbose samples can prompt MLLMs to generate sequences that include objects not present in the visual input, thereby leading to longer sequences and higher energy-latency cost. 

\textbf{Visual Interpretation.} 
We adopt GradCAM \cite{selvaraju2017grad}, a gradient-based visualization technique that generates attention maps highlighting the relevant regions in the visual input for the generated sequences. From Fig. \ref{grad cam image} and Fig. \ref{grad cam video}, the attention of original images and videos primarily concentrates on a local region containing a specific object mentioned in the generated caption. In contrast, our verbose images can effectively disperse attention and cause  image-based LLMs to shift their focus from a specific object to the entire image region.  Similarly, our verbose videos are shown to distribute attention throughout the video frames in both spatial and temporal dimensions, rather than focus on specific actions or scenes. Since the attention mechanism serves as a bridge between the visual input and the output sequence of MLLMs, we conjecture that the generation of a longer sequence can be reflected in an inaccurate focus and dispersed and uniform attention from the visual input.

In summary, our proposed verbose samples exhibit a higher hallucination rate and uniform attention distribution. Therefore, we hope that our proposed unified interpretation will serve as a valuable analytical tool for further research.

\subsection{Discussions}
\textbf{Different perturbation magnitudes.} In our default setting, the perturbation magnitude $\epsilon$ is set as 8. To investigate the impact of different magnitudes, we vary $\epsilon$ under $[2,4,8,16,32]$ in Table \ref{tab:varying perturbation magnitude} and calculate the corresponding LIPIS \cite{zhang2018unreasonable} between  original ones and their verbose counterpart, which quantifies the perceptual difference. It can be observed that a larger perturbation magnitude $\epsilon$ results in a longer generated sequence by MLLMs but produces more perceptible verbose samples. Consequently, this trade-off between visual perceptibility and energy-latency cost highlights the importance of choosing an appropriate perturbation magnitude during evaluation. 

\begin{figure*}[t] \centering   
\begin{minipage}{\textwidth}
    \includegraphics[width=\columnwidth]{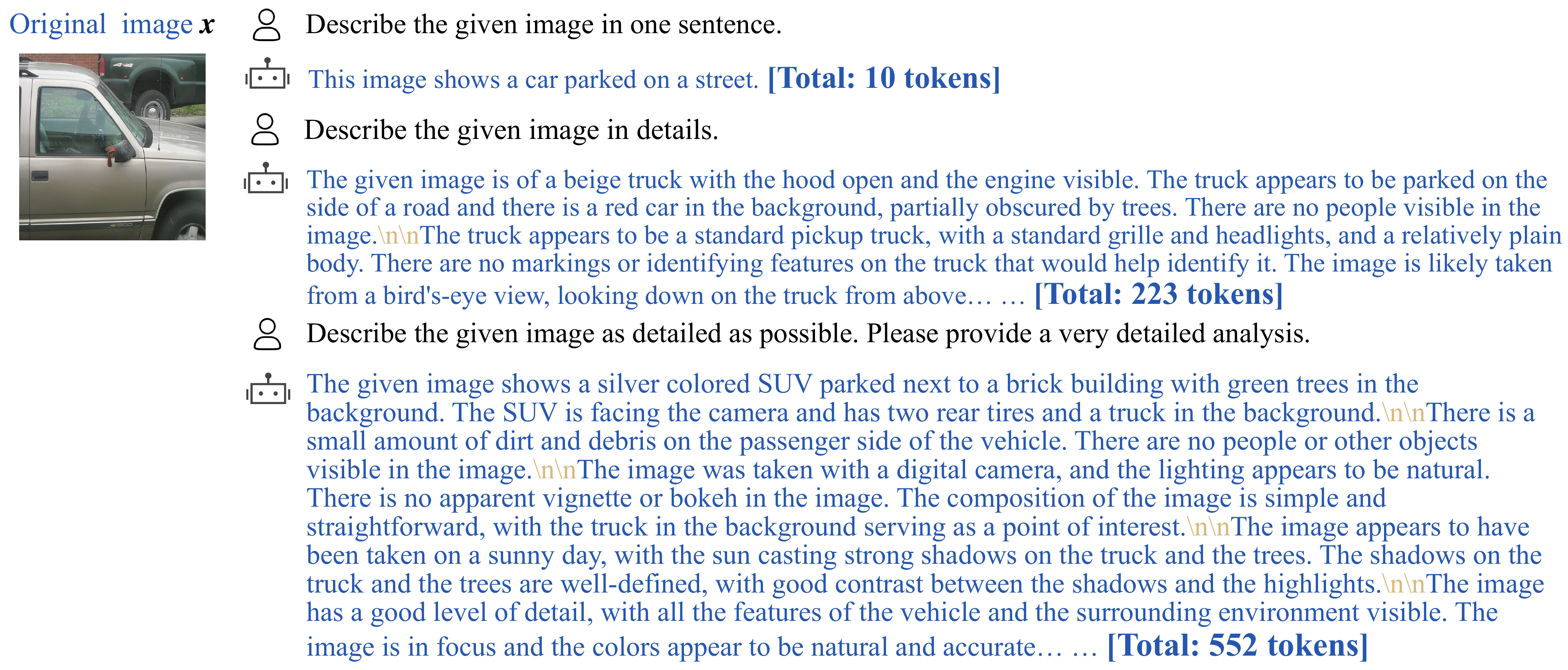} 
\end{minipage}
\caption{An example of generated sequences from MiniGPT-4 by different input prompts. Users have diverse requirements and input data, leading to a wide range of lengths of generated sequences.}
\vspace{-1.5em}
\label{different text prompts introduce different generation length}
\end{figure*}

\begin{table*}[t]
\caption{The length of generated sequences, energy consumption (J), and latency time (s) against BLIP-2 and VideoChat-2 on captioning and question answering tasks. Our verbose samples can still achieve better on different multi-modal tasks.}
\vspace{-1em}
\label{tab:different tasks on vqa and reason}
\centering
\small
\setlength\tabcolsep{2.5pt}{
\begin{tabular}{@{}l|ccc|ccc|ccc|ccc@{}}
\toprule
\multirow{2}{*}{Attacking method} & \multicolumn{3}{c|}{Image Caption} & \multicolumn{3}{c|}{Image QA} & \multicolumn{3}{c|}{Video Caption} & \multicolumn{3}{c}{Video QA}\\
 & Length & Latency & Energy & Length & Latency & Energy & Length & Latency & Energy & Length & Latency & Energy \\
\midrule 
Original & 8.82 & 0.39 & 16.08 & 6.43 & 0.44 & 17.92 & 13.97 & 4.79 & 153.08 & 16.66 & 5.67 & 181.13 \\
Noise & 9.55 & 0.43 & 17.53 & 6.62 & 0.45 & 17.09 & 14.49 & 4.84 & 156.84 & 16.94 & 7.04 & 211.58 \\
Sponge samples & 22.53 & 0.73 & 30.20 & 133.24 & 5.04 & 191.16 & 28.84 & 6.99 & 222.83 & 22.89 & 5.91 & 170.45 \\
NICGSlowDown & 103.54 & 3.78 & 156.61 & 127.96 & 5.07 & 190.13 & 29.99 & 7.02 & 224.11 & 23.69 & 6.41 & 201.94 \\
\textbf{Verbose samples (Ours)} & \textbf{226.72} & \textbf{7.97} & \textbf{321.59} & \textbf{271.95} & \textbf{11.49} & \textbf{365.49} & \textbf{165.77} & \textbf{35.88} & \textbf{1323.42} & \textbf{112.09} & \textbf{24.11} & \textbf{864.65} \\
\bottomrule
\end{tabular}}
\end{table*}

\textbf{Different adversary knowledge.} We explore a different adversary knowledge from main experiments, where the victim MLLMs are unknown in black-box settings \cite{ilyas2018black,bai2020improving}. To induce high energy-latency cost of black-box MLLMs, we can leverage the transferability property \cite{dong2018boosting} of our verbose samples. We can first craft verbose samples on a known and accessible surrogate model and then utilize them to transfer to the target victim MLLMs. The black-box transferability results across four image-based LLMs of our verbose images are evaluated in Table \ref{tab:transferability of four models of our verbose images}, while the transferability results of our verbose videos are assessed across three video-based LLMs, presented in Table \ref{tab:transferability of four models of our verbose images of videos}. When the source model is set as `None', it indicates that we evaluate the energy-latency cost for the target model by utilizing the original samples. The results demonstrate that our transferable verbose samples are less effective than the white-box verbose samples but still result in a longer generated sequence.

\textbf{Different multi-modal tasks.} To verify the effectiveness of our methods, we induce high energy-latency cost on an additional multi-modal task: visual question answering (VQA). For verbose images, we use VQAv2 dataset \cite{goyal2017making} in image question answering and for verbose videos, we employ MSVD-QA dataset \cite{chen2011collecting} in video question answering. We use BLIP-2 as the target image-based LLM and VideoChat-2 as the target video-based LLM. Unless otherwise specified, other settings remain unchanged. Table \ref{tab:different tasks on vqa and reason} demonstrates that our verbose samples can induce the highest energy-latency cost among different multi-modal tasks.

\textbf{Potential defenses.} An intuitive defense to mitigate the energy-latency vulnerability is to impose a limitation on the length of generated sequences. We argue that such an intuitive defense is infeasible and the reason is as follows. 

(1) Users have diverse requirements and input data, leading to a wide range of sentence lengths and complexities. For example, the prompt text of `Describe the given image in one sentence.' and `Describe the given image in details.' can introduce different lengths of generated sentences. We visualize a case in Fig. \ref{different text prompts introduce different generation length}. Consequently, service providers often consider a large token limit to accommodate these diverse requirements and ensure that the generated sentences are complete and meet users' expectations. Previous work, NICGSlowDown \cite{chen2022nicgslowdown}, also states the same view as us. Besides, as shown in Table \ref{tab:different tasks on vqa and reason}, the results can demonstrate that our verbose samples are adaptable for different prompt texts and can induce the length of generated sentences closer to the token limit set by the service provider. As a result, the energy-latency cost can be increased while staying within the imposed constraints.

(2) We argue that this attack surface about availability of MLLMs becomes more important and our verbose samples can induce more serious attack consequences in the era of (multi-modal) large language models. The development of MLLMs has led to models capable of generating longer sentences with logic and coherence. Consequently, service providers have been increasing the maximum allowed length of generated sequences to ensure high-quality user experiences. For instance, gpt-3.5-turbo and gpt-4-turbo allow up to 4,096 and 8,192 tokens, respectively. Hence, we would like to uncover that while longer generated sequences can indeed improve service quality, they also introduce potential security risks about energy-latency cost, as our verbose samples demonstrate. Therefore, when MLLMs service providers consider increasing the maximum length of generated sequences for better user experience, they should not only focus on the ability of MLLMs but also take the maximum energy consumption payload into account.  

In terms of potential defenses, common data pre-processing methods, such as compressing or generative models, should be beneficial in defending our verbose samples. Furthermore, it is crucial to focus on safety-based alignment strategies, as they can significantly enhance the robustness of MLLMs against attacks. In our future work, we plan to investigate and develop defense methods to address the security threat introduced by energy-latency manipulation in a comprehensive manner.

\section{Conclusion and limitation}
In this paper, we aim to craft an imperceptible perturbation to induce high energy-latency cost of MLLMs during the inference stage. We propose verbose samples to prompt MLLMs to generate as many tokens as possible. 
Extensive experimental results demonstrate that, compared to original images, our verbose images can increase the average length of generated sequences by 7.87$\times$ and 8.56$\times$ on MS-COCO and ImageNet across four image-based LLMs. Our verbose videos can increase the average length of generated sequences by 4.04$\times$ and 4.14$\times$ compared to original videos on MSVD and TGIF datasets across three video-based LLMs. In addition, we compare and discuss differences between verbose images and videos, highlighting the significance of the proposed verbose videos for video-based LLMs instead of directly using a single verbose image or a combination of multiple verbose images.
For a deeper understanding, we propose a unified framework from hallucination and visual attention to interpret the mechanism behind the energy-latency manipulation of MLLMs.
We hope that our verbose samples can serve as a baseline to manipulate high energy-latency cost of MLLMs.

The primary limitation of our verbose samples lies in the focus on the digital world, where input samples are directly fed into the models. As technology progresses, MLLMs are expected to be deployed in more complex, real-world scenarios, such as autonomous driving. In these situations, input samples would not be pre-recorded but captured in real-time from physical environments using cameras. Therefore, the current verbose samples may not fully address the challenges and complexities associated with energy-latency manipulation in real-world, dynamic settings. Future research should investigate the implementation and impact of energy-latency manipulation techniques in the physical world to ensure their effectiveness in practical applications.

Please note that we restrict all experiments in the laboratory environment and do not support our verbose samples in the real scenario. The purpose of our work is to raise the awareness of the security concern in availability of MLLMs and call for  practitioners to pay more attention to the manipulation of energy-latency cost for MLLMs and model trustworthy deployment.

\ifCLASSOPTIONcaptionsoff
  \newpage
\fi



%


\bibliographystyle{IEEEtran}
\bibliography{egbib}

%


\end{document}